\relax
\documentclass[letterpaper]{article} 
\usepackage{algorithm}
\usepackage{algpseudocode}
\usepackage{aaai19}  
\usepackage{times}  
\usepackage{helvet}  
\usepackage{courier}  
\usepackage{url}  
\usepackage{graphicx}  
\usepackage{hyperref}
\usepackage{amsmath,epsfig}
\usepackage{nccmath}
\usepackage{stfloats}
\usepackage{multirow}
\usepackage{amssymb}
\usepackage{epstopdf}
\usepackage{enumitem}
\usepackage{mathrsfs}
\usepackage{cleveref}
\usepackage{array}
\usepackage{amsthm}
\usepackage{float}
\usepackage{array}
\usepackage{color,soul}

\newcommand{\RNum}[1]{\uppercase\expandafter{\romannumeral #1\relax}}
\newtheoremstyle{definitionAnh}
  {14pt} 
  {14pt} 
  {} 
  {} 
  {\bfseries} 
  {} 
  {.5em} 
  {} 
\makeatletter
\g@addto@macro\th@definition{\thm@headpunct{}}
\makeatother
\theoremstyle{definitionAnh}
\newtheorem{theorem}{Theorem}

\newtheorem{proposition}[theorem]{Proposition}
\newtheorem{lemma}[theorem]{Lemma}

\begin{document}
\title{Data Masking with Privacy Guarantees}

 \author{Anh T. Pham \\
	Oregon State University\\phamtheanhbka@gmail.com \And Shalini Ghosh \\Samsung Research\\shalini.ghosh@gmail.com  \And Vinod Yegneswaran\\
 	SRI international\\vinod@csl.sri.com}

\maketitle
\begin{abstract}
We study the problem of data release with privacy, where data is made
available with privacy guarantees while keeping the usability of the
data as high as possible --- this is important in health-care and
other domains with sensitive data. In particular, we propose a method
of masking the private data with privacy guarantee while ensuring that a
classifier trained on the masked data is similar to the classifier
trained on the original data, to maintain usability. We analyze the theoretical risks of the proposed method and the traditional input perturbation method. Results show that the proposed method achieves lower risk
compared to the input perturbation, especially
when the number of training samples gets large. 
We illustrate the
effectiveness of the proposed method of data masking for
privacy-sensitive learning on $12$ benchmark datasets.
\end{abstract}

\section{Introduction}
\label{sec:intro}

In domains like healthcare or finance, data can be sensitive and
private. There are several scenarios where a dataset needs to be
shared while protecting sensitive parts of the data. For example,
consider a medical study where a group of patients with a particular
medical condition are being studied. The identifying data of some
patients (e.g., those with a rare disease) may need to be masked while
sharing their records with a wider group of medical
researchers. However, when the patient records are processed by
clinical decision support tools, we want the machine learning (ML)
models in the tools to have similar performance on the masked data as
they would on the original data.

Several approaches have been proposed to preserve privacy of data,
e.g., by anonymization~\cite{samarati1998generalizing}, by
generalization \cite{mohammed2011differentially}. Methods for differential-privacy include adding
Laplace-noise~\cite{sarwate2013signal}, modifying the objective \cite{chaudhuri2009privacy}, and posterior sampling
\cite{dimitrakakis2014robust,wang2015privacy}.  Privacy-preserving data publishing transforms
sensitive data to protect it against privacy attacks while supporting
effective data mining tasks~\cite{fung2010privacy}.  Differentially
private data release~\cite{mohammed2011differentially} presents an
anonymization algorithm that satisfies the $\epsilon-$differential
privacy model, while other methods of data
release~\cite{chen2011publishing,xiao2010differentially} group the
data and add noise to the partition counts. However, these techniques
don't explicitly try to maintain the accuracy of a model. Our approach
masks training samples with less sensitive ones with privacy guarantee, while
ensuring that the classifier trained on the masked data
reaches accuracy similar to the classifier trained on the original data. Moreover, compared to publishing masked classifier, publishing masked data enables
other types of classifiers to be trained by the user. There are also query-based data masking methods for a classifier, which are sparse vector techniques for generating masked data using a query that the gradient of the masked data is zero \cite{dwork2014algorithmic,lyu2017understanding,lee2014top,blum2008learning}. However, when the gradient computation is complicated, designing a method to achieve a zero gradient can be tricky.

We have three main contributions in this paper. First, we propose a
novel algorithm of data masking for privacy-sensitive
learning. Second, we provide a theoretical guarantee explaining why
the proposed method is more suitable for a large number of training
samples than a traditional input perturbation method. Finally, we
illustrate the efficacy of our method considering logistic regression
as an example classifier, on both synthetic and $12$ benchmark datasets.

\section{Problem setting}
\label{sec:problem}

\noindent{\bf Goal:} Assume we train a model parameterized by ${\bf w}
\in \mathbb{R}^d$ on a dataset $\mathbb{D}_{train}=\{{\bf
  x}_i,y_i\}_{i=1}^{N}$, where ${\bf x}_i \in \mathbb{X}=
\mathbb{R}^d$, $y_i \in \mathbb{Y}=\{-1, 1\}$, and $d$ is the number
of features. The goal of our data publishing algorithm $\mathbb{A}$ is
generating a masked training dataset $\mathbb{D}_{masked}=\{{\bf
  x}'_i,y_i\}_{i=1}^{N}$, where ${\bf x}'_i \in \mathbb{X}$,  such that: (a) $\mathbb{D}_{masked}$ is as different
as possible from $\mathbb{D}_{train}$, but (b) the model trained on
$\mathbb{D}_{masked}$ gives us parameters ${\bf w}'$ that are close to
the original parameters ${\bf w}$ of the model trained on
$\mathbb{D}_{train}$.

This paper outlines an approach for achieving this goal. Before that,
we review several concepts of data publishing with privacy and the core
formulation of logistic regression.

\subsection{Data publishing with differential privacy (DPDP)}
\label{sec:problem:data}

We first begin with the concept of data publishing with differential
privacy (DPDP).
We consider two datasets of $N$ training samples,
$\mathbb{D}_{train1}=\{{\bf x}_{1i},y_{1i}\}_{i=1}^N$ and
$\mathbb{D}_{train2}=\{{\bf x}_{2i},y_{2i}\}_{i=1}^N$, which are
different at only one sample: without loss of generality, assume
${\bf x}_{1i}={\bf x}_{2i}$ and $y_{1i}=y_{2i}$ for $i=\{1,
2,\dots,N-1\}$, and ${\bf x}_{1N}\neq {\bf x}_{2N}$ and (or) $y_{1N} \neq
y_{2N}$. A data publishing algorithm $\mathbb{A}$ is said to be
$\epsilon$-private \cite{dwork2008differential} if
\begin{eqnarray}
\frac{p\big(\mathbb{A}(\mathbb{D}_{train1})=\mathbb{O}\big)}{p(\mathbb{A}\big(\mathbb{D}_{train2})=\mathbb{O}\big)} <e^{\epsilon},\nonumber
\end{eqnarray}
where $\mathbb{O}=\{{\bf x}'_i,y_i\}_{i=1}^{N}$ is a
particular output of the data publishing algorithm
$\mathbb{A}$. Intuitively, differential privacy guarantees that for small
$\epsilon$, the output of $\mathbb{A}$ is not sensitive to the
existence of a single sample in the dataset. In this setting, the
attacker has less chance to infer details about a particular training
sample in the data. In this work we focus on differential privacy for
masked data generation where the machine learning algorithm we consider
is logistic regression~\cite{walker1967estimation}.

\subsection{Core formulation of logistic regression}
\noindent {\bf Logistic Regression:} We are given a training dataset
$\mathbb{D}_{train}$. The goal for training a logistic regression
classifier is finding a mapping function between a sample in $\mathbb{R}^d$ and
a label in $\{1,-1\}$. Specifically, we model the relation among
a sample ${\bf x}_i$ and its label $y_i$ as
\begin{equation}
p(y_i|{\bf x}_i,{\bf w})=\frac{e^{y_i{\bf w}^T{\bf x}_i}}{1+e^{y_i{\bf w}^T{\bf x}_i}}.\nonumber
\end{equation}
Assuming samples are i.i.d., the log-likelihood for the training samples is
\begin{equation}
\mathbb{L}_{\lambda}({\bf w})=\frac{1}{N}\sum_{i=1}^{N} \Big[\big(y_i{\bf w}^T{\bf x}_i\big) -\log(1+ e^{y_i{\bf w}^T{\bf x}_i})\Big]+\lambda \frac{||{\bf w}||^{2}}{2},
\label{e:LR_train}
\end{equation}
with $\lambda$ is the regularization parameter and $||\cdot||$ denotes
the 2-norm.

\noindent {\bf Training logistic regression:} In logistic regression,
training is done by finding the parameter ${\bf w}$ that maximizes the
log-likelihood in \eqref{e:LR_train}, i.e., the gradient of $\mathbb{L}_{\lambda}$ at
${\bf w}$ is $0$, as follows:
\begin{equation}
\frac{\partial \mathbb{L}_{\lambda}({\bf w})}{\partial {\bf w}}=\frac{1}{N}\Big[\sum_{i=1}^{N}\Big[y_{i}-p(y_{i}=1|{\bf x}_i,{\bf w})\Big]{\bf x}_i\Big] +\lambda {\bf w}=0.
\label{e:real}
\end{equation}
For various logistic regression optimization techniques to make the above gradient $0$, please refer to \cite{minkacomparison}.

\section{DPDP by Masked Data Generation}
\label{sec:dpdp:masked}

In this section, we describe how to generate masked samples for
logistic regression.\footnote{In this work we consider logistic regression as the classifier. The work flow of data publishing for other classifiers, e.g., SVM, is similar to that of the proposed method.}

\subsection{Adding Laplace noise to the classifier}
\begin{figure}
\centerline{\includegraphics[width=3.7in]{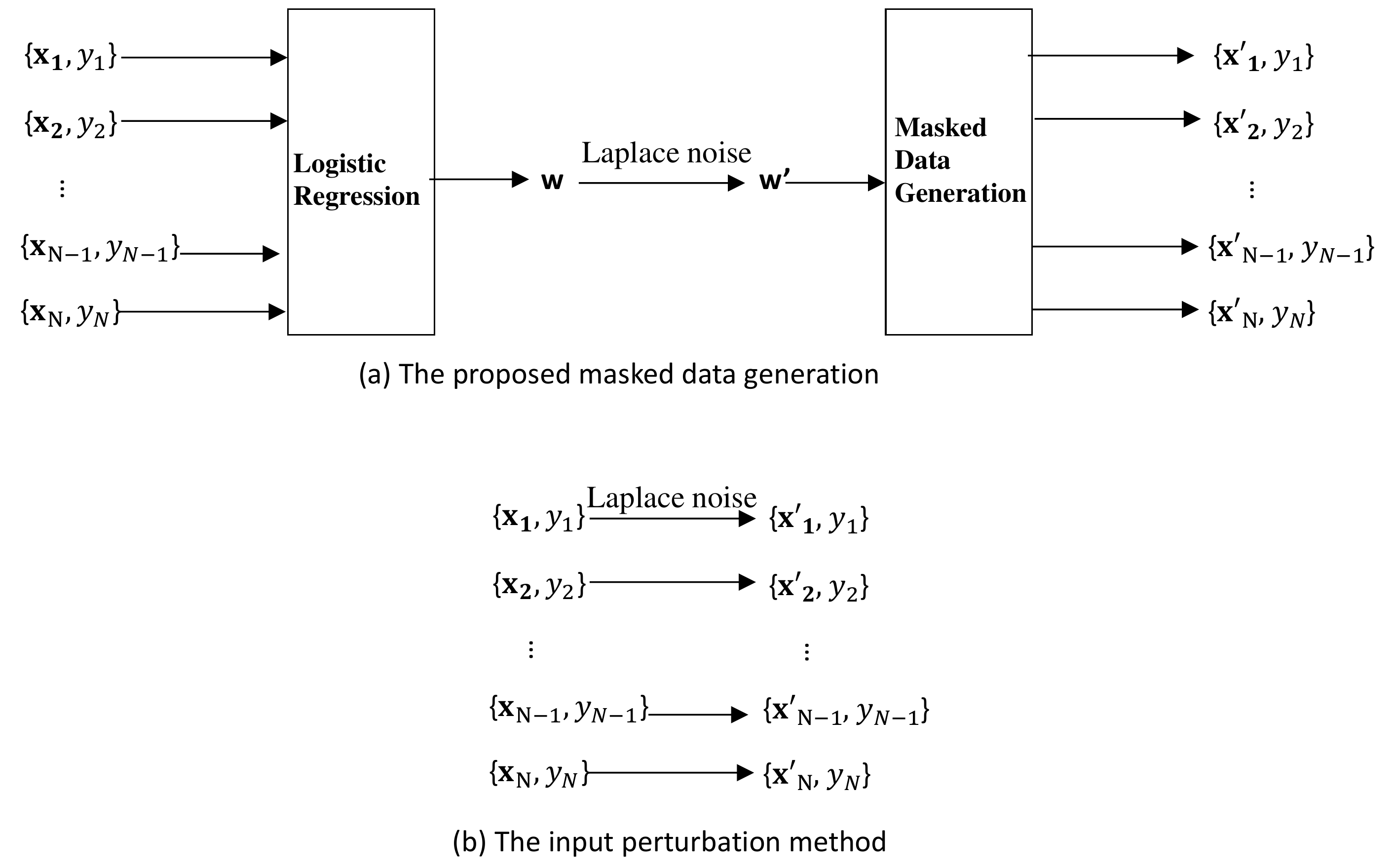}}
\caption{The work flow of the proposed method and traditional data publishing methods.}
\label{fig:work_flow}
\end{figure}

Unlike previous approaches of adding noise to the data then
publishing noisy data, we consider a novel approach: we first train
a classifier on the original data, and then add Laplace noise to the
classifier. The motivation for adding noise is that in differential
privacy, the goal is to make similar output for any two neighbor
datasets $\mathbb{D}_1$ and $\mathbb{D}_2$ so that attacker cannot
infer about the existence of any single training sample. Since the
classifiers trained on two datasets $\mathbb{D}_1$ and $\mathbb{D}_2$
are not equal, adding Laplace noise to the parameters of those
classifiers would account for that difference, and with some
probability those classifiers after adding noise would be
equal. Subsequently, we generate and publish a masked dataset such that the
gradient of the log-likelihood for the noisy classifier is $0$. The
work flow of the proposed framework is illustrated in
Fig.~\ref{fig:work_flow}(a). In comparison, the work flow of traditional
data publishing methods by perturbation is shown in
Fig.~\ref{fig:work_flow}(b).

\subsection{Generating masked data}
We generate masked data $\mathbb{O}=\{{\bf x}'_i\}_{i=1}^N$ such that the gradient of the log-likelihood of $\mathbb{O}$ for the
aforementioned noisy classifier ${\bf w}'$ is $0$. The
optimal condition for masked data is the following:
\begin{equation}
\frac{1}{N}\Big[\sum_{i=1}^{N}\Big[y_{i}-p(y_{i}=1|{\bf x}'_i,{\bf w}')\Big]{\bf x}'_i\Big] +\lambda {\bf w}'=0,
\label{e:masked}
\end{equation}
\noindent where the masked samples $\{{\bf x}'_i, y'_i\}$(s) are unknown. To evaluate the optimality of the set $\mathbb{S}$ of masked samples w.r.t.~${\bf w}'$, we use the 2-norm of the gradient:
\begin{eqnarray}
  &&\mathcal{N}(\mathbb{S})=\Big|\Big|\sum_{i=1}^{N}\Big(y_{i}-p(y_{i}=1|{\bf x}'_i,{\bf w}'){\bf x}'_i\Big)+N\lambda {\bf w}'\Big|\Big|^2 \nonumber
\end{eqnarray}
We start with an initial set of training samples
$\mathbb{S}$ then iteratively add new sample to
$\mathbb{S}$. The criteria to evaluate the new sample is the 2-norm
of the gradient of $\mathbb{S}$ after including the new sample.

Algorithm \ref{a:DFG} outlines our proposed Masked Data Generation
algorithm. The algorithm terminates when the number of samples in $\mathbb{S}$ reaches $N$.

\renewcommand{\algorithmicrequire}{\textbf{Input:}}
\renewcommand{\algorithmicensure}{\textbf{Output:}}
\newcommand{\algrule}[1][.2pt]{\par\vskip.5\baselineskip\hrule height #1\par\vskip.5\baselineskip}
\begin{algorithm}[H]
\caption{Masked Data Generation}
\begin{algorithmic}
\Require $N$ training samples $\mathbb{D}_{train}=\{{\bf x}_i\}_{i=1}^N$, $\epsilon$
\Ensure $N$ masked training samples $\mathbb{O}=\{{\bf x}'_i\}_{i=1}^N$.
\\\algrule
{\bf Step 1:} Train Logistic regression classifier ${\bf w}$, as in \eqref{e:real}.\\
{\bf Step 2:} Add Laplace noise to the classifier ${\bf w}'={\bf w}+\eta$, where $||\eta|| $c$ \sim e^{-\frac{\lambda N \epsilon}{2}||\eta||}$, where $c$ is a normalized constant.\\
{\bf Step 3:} $\mathbb{S}=\{\emptyset\}$. Incrementally generate masked samples.
\While{$cardinality(\mathbb{S})\le N$}\\
Find an outliers $\{{\bf x}'_m\}$ reducing the 2-norm of the gradient of $\mathbb{S}$ the most, using Gradient Descent \eqref{e:grad}\\
Add the new sample $\mathbb{S}=\mathbb{S} \bigcup \{{\bf x}'_m\}$
\EndWhile\\
Return $\mathbb{O}=\mathbb{S}$;
\end{algorithmic}
\label{a:DFG}
\end{algorithm}

\subsection{Iteratively generating masked samples}
In this section, we present the gradient descent method to iteratively
generate masked samples. In particular, given the current set of
masked samples $\mathbb{S}$, we need to find the next masked sample
$\{{\bf x}'_m\}$ such that the 2-norm of the gradient of the set
$\mathbb{S} \bigcup \{{\bf x}'_m\}$ is close to $0$ as possible.

For simplicity of notation, denote $\sum_{{\bf x}'_i \in
  \mathbb{S}}\Big(y_{i}-p(y_{i}=1|{\bf x}'_i,{\bf w}){\bf
  x}'_i\Big)+N\lambda {\bf w}=g$ as the current gradient of the
current masked samples. Consequently, we need to find the next masked
sample $\{{\bf x}'_m\}$ minimizing the following objective
\begin{eqnarray}
\mathcal{N}({\bf x}'_m)=||\big(y_m&-&\frac{e^{y_m{\bf w}^T {\bf x}'_m}}{1+e^{{\bf w}^T {\bf x}'_m}}\big){\bf x}'_m +g ||^2.
\label{e:quadratic}
\end{eqnarray}
To minimize  \eqref{e:quadratic}, we use backtracking gradient descent. The gradient is computed as
\begin{align}
&\frac{\partial \mathcal{N} ({\bf x}'_m)}{ \partial {\bf x}'_m}=2\Big(\big(y_m-\frac{e^{y_m{\bf w}^T {\bf x}'_m}}{1+e^{{\bf w}^T {\bf x}'_m}}\big){\bf x}'_m +g\Big)\times\nonumber\\
&\Big((y_m-\frac{e^{{\bf w}^T {\bf x}'_m}}{1+e^{{\bf w}^T {\bf x}'_m}})\mathbb{I}-{\bf w}^T{\bf x}'_m\frac{e^{{\bf w}^T {\bf x}'_m}}{(1+e^{{\bf w}^T {\bf x}'_m})^2}\Big),
\label{e:grad}
\end{align}
where $\mathbb{I}$ is the identity matrix in $\mathbb{R}^{d\times
  d}$. Note that, we can generalize our algorithm to $C$ classes, with
$C>2$, as follows
\begin{eqnarray}
&&\frac{\partial \mathcal{N}}{ \partial {\bf x}'_m}=\sum_{c=1}^{\mathcal{C}}2\Big([I(y_m=c)-p(y_m=c|{\bf x}'_m,{\bf w})]{\bf x}'_m +g_c\Big)\times\nonumber\\
&&\Big([I(y_m=c)-p(y_m=c|{\bf x}'_m,{\bf w})]-\nonumber\\
&&\sum_{l=1}^C [p(y_m=c|{\bf x}'_m,{\bf w})p(y_m=l|{\bf x}'_m,{\bf w})({\bf w}_c-{\bf w}_l)^T{\bf x}'_m)]\Big).\nonumber
\end{eqnarray}
\noindent {\bf Computational complexity:} The computational complexity of the proposed algorithm is linear in term of number of added samples.

\noindent {\bf Intuition: } Most differential privacy algorithms for
data publishing modify the data by adding uniform noise, e.g., as in Fig.~\ref{fig:work_flow}(b), which may
change the original data manifold closer to a uniform manifold and may not be optimized for any particular machine learning model.  

{\bf Comparison to classifier publishing:} The proposed approach has an advantage over other traditional approaches. In particular, assuming a non-empty initialized set of training samples $\mathbb{S}$ in Step $3$ of Algorithm \ref{a:DFG}, the proposed method adds fake samples with completely different manifold
to the dataset. For example, assume
we want to preserve the privacy of a dataset consisting of
non-diabetes patients and {\em sensitive type-1 diabetes patients}. We can initially
add {\em non-sensitive type-2 diabetes} data samples, thereby preserving the
privacy of the type-1 diabetes patients. Moreover, by iteratively adding masked samples, a classifier that is
trained on the original data will be quite close to the classifier
trained on the new masked data. Compared to  publishing the noisy classifier as in \cite{chaudhuri2009privacy}, the proposed data masking method allows users to benefit from real data, i.e., in this case non-diabetes and type-2 diabetes data, and train other types of classifiers on them.

\subsection{Privacy guarantee of Masked Data Generation}
There are two aspects of a data publishing algorithm. First, we need
to guarantee that the algorithm is $\epsilon$-private. In particular,
is the algorithm sensitive to the existence of a single sample in two
datasets that are different only at that sample? Second, we would
like to assess how the utility of the published dataset changes with
changing $\epsilon$. The following Proposition answers the first question.

\begin{proposition}If $||{\bf x}_i|| \le 1, \forall i$, then Algorithm \ref{a:DFG} is $\epsilon$-private.
\label{l:4}
\end{proposition}

\subsection{Utility of Masked Data Generation with changing $\epsilon$}
\label{ss:u_MDG}
We next consider the utility aspect of the masked dataset $\mathbb{O}$
with different values of $\epsilon$.  We consider the utility of the
published data to be how well the classifier trained on the published
data is close to the classifier trained on the original data. 

Let us suppose that training logistic regression on the original
dataset $\mathbb{D}_{train}$ and the masked dataset $\mathbb{O}$ gives
us parameters ${\bf w}$ and ${\bf w}'$, respectively. We are
interested in comparing the $0/1$ risk
\cite{vapnik1998statistical} of the classifier
trained on masked data (${\bf w}'$), to the $0/1$ risk of the
classifier trained on original data (${\bf w}$).  Note that
logistic regression is classification calibrated \cite{bartlett2006convexity}, which means that
minimizing the negative log-likelihood leads to minimizing the $0/1$
risk. Thus, it is sufficient to compare the log-likelihood 
$\mathbb{L}_{\lambda}$ of ${\bf w}'$
compared to that of ${\bf w}$.

\begin{proposition} With probability $1-\delta$, $\mathbb{L}_{\lambda}({\bf w}')-\mathbb{L}_{\lambda}({\bf w})\le\frac{1}{2}(\frac{2 d \log (\frac{d}{\delta})}{\lambda N\epsilon})^2(\lambda +1)$.
\label{l:1}
\end{proposition}

From Lemma \ref{l:1}, the classifier trained on masked data improves
when $N$ is larger.

\section{DPDP by Input Perturbation}
\label{sec:dpdp:perturb}

In this section, we consider a classical and natural algorithm to
publish data \cite{sarwate2013signal,mivule2012utilizing}. The
algorithm is quite simple: it directly adds noise $\eta \sim
e^{\frac{-\epsilon||\eta||}{2}}$ to each input sample. The detailed
algorithm is shown in Algorithm \ref{a:IPB}.  Similar to Algorithm
\ref{a:DFG}, in the rest of this section we consider the privacy and
the utility of the input perturbation algorithm when $\epsilon$
changes.

\subsection{Privacy guarantee of Input Perturbation}

We first show that Algorithm \ref{a:IPB} is $\epsilon$-private.

\begin{proposition} If $||{\bf x}_i|| \le 1, \forall i$, then algorithm \ref{a:IPB} is $\epsilon$-private.
\label{l:40}
\end{proposition}

\begin{algorithm}[H]
\caption{Input Perturbation}
\begin{algorithmic}
\Require $N$ training samples $\mathbb{D}_{train}=\{{\bf x}_k\}_{k=1}^N$, $\epsilon$
\Ensure $N$ masked training samples $\mathbb{O}=\{{\bf x}'_k\}_{k=1}^N$
\\\algrule
\While{$k<N$}\\
$\eta \sim e^{\frac{-\epsilon||\eta||}{2}}$\\
${\bf x}_k'={\bf x}_k+\eta$\\
$k=k+1$
\EndWhile\\
Return $\mathbb{O}=\{{\bf x}_k'\}_{k=1}^N$\\
\end{algorithmic}
\label{a:IPB}
\end{algorithm}

\subsection{Utility of Input Perturbation with changing $\epsilon$}

Similar to Section \ref{ss:u_MDG}, we consider the log-likelihood of the classifier ${\bf
  w}'$ trained on perturbed data. We are going to bound
the log-likelihood w.r.t.~the original data $\mathbb{L}_{\lambda}({\bf
  w}')-\mathbb{L}_{\lambda}({\bf w})$. We begin with the
following Proposition.

\begin{lemma}~\cite{chaudhuri2009privacy}.
Let $G({\bf w})$ be a convex function and $g({\bf w})$ be a function
with $||\nabla g({\bf w})|| \le g_1$ and $\min_v \min_{{\bf
    w}}||v^T\nabla^2 (G+g)({\bf w})v|| \ge G_2$. Let ${\bf
  w}=\arg\min G({\bf w})$ and ${\bf w}'=\arg\min G({\bf w}) +
g({\bf w})$. Then $||{\bf w}'-{\bf w}|| \le \frac{g_1}{G_2}$.
\label{l:2}
\end{lemma}

\begin{proposition}
With probability $1-\delta$, $\mathbb{L}_{\lambda}({\bf
  w}')-\mathbb{L}_{\lambda}({\bf w}) \le \frac{1}{2}(\frac{2 d
  \log \frac{d}{\delta}}{\lambda \epsilon})^2(\lambda+1)$.
\label{l:3}
\end{proposition}

\noindent From Proposition~\ref{l:3}, the classifier trained on perturbed data does
not improve when $N$ is larger, as we see in Proposition~\ref{l:1}.

\section{Experiments}
\label{sec:expt}

We compare the performance of our Masked Data Generation method in
Algorithm~\ref{a:DFG} to the Input Perturbation method in
Algorithm~\ref{a:IPB}, on both synthetic and real datasets.
\begin{figure*}[hbtp]
\begin{minipage}[b]{0.31\linewidth}
  \centering
  \centerline{\includegraphics[width=3.5cm]{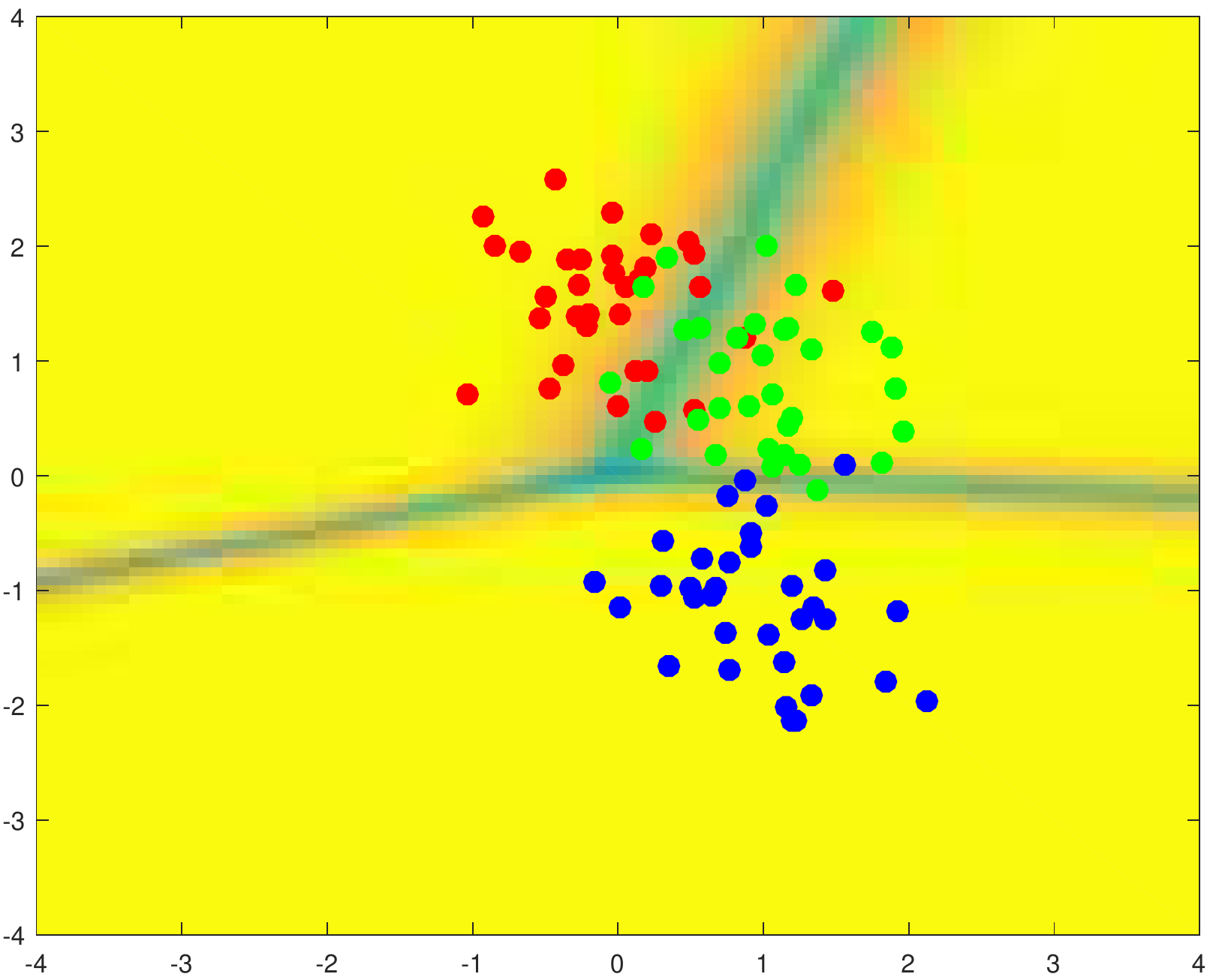}}
  \centerline{(a) True training samples}\medskip
\end{minipage}
\hfill
\begin{minipage}[b]{0.31\linewidth}
  \centering
  \centerline{\includegraphics[width=3.5cm]{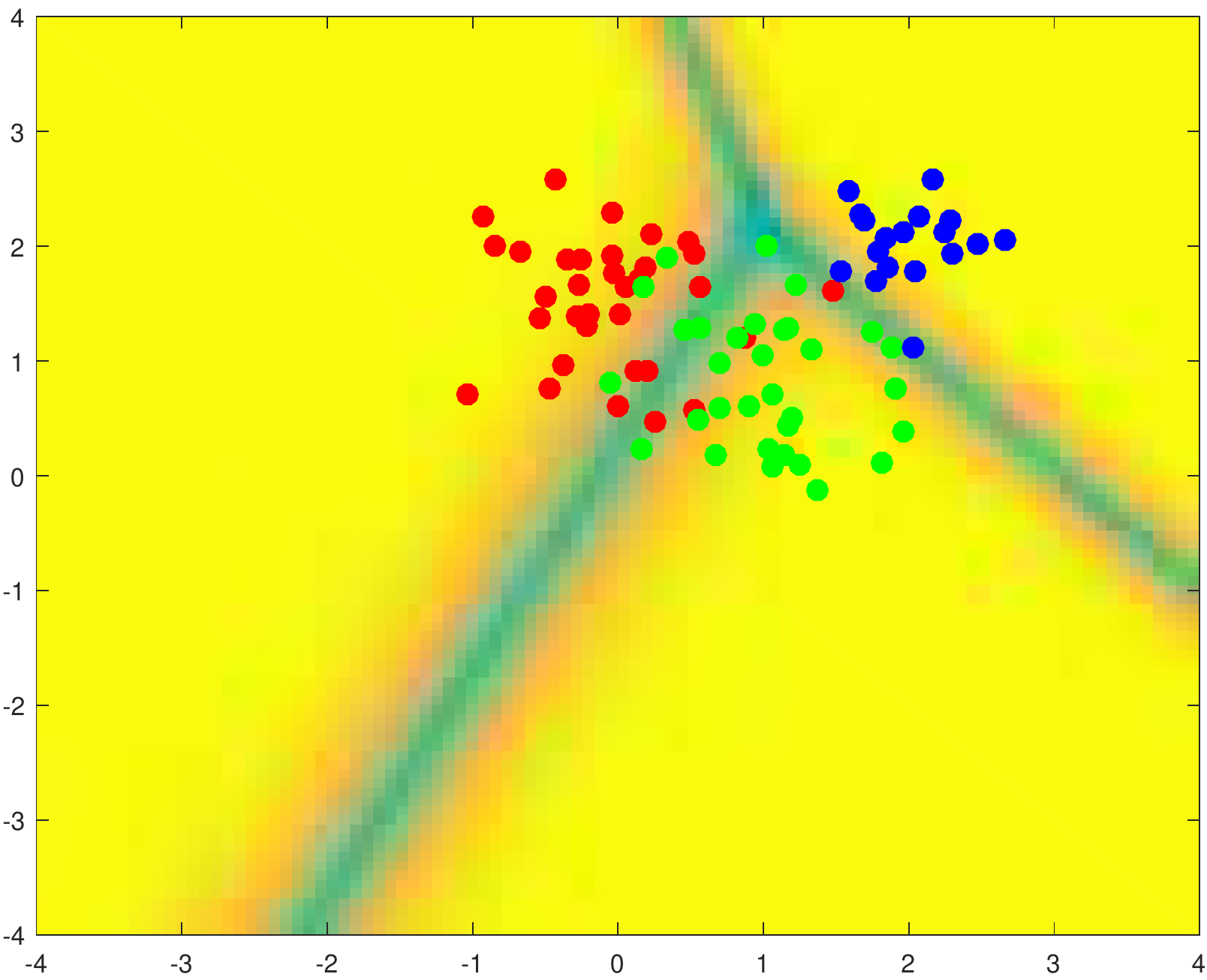}}
  \centerline{(b) Initially masked samples set $\mathbb{S}$}\medskip
\end{minipage}
\hfill
\begin{minipage}[b]{0.31\linewidth}
  \centering
  \centerline{\includegraphics[width=3.5cm]{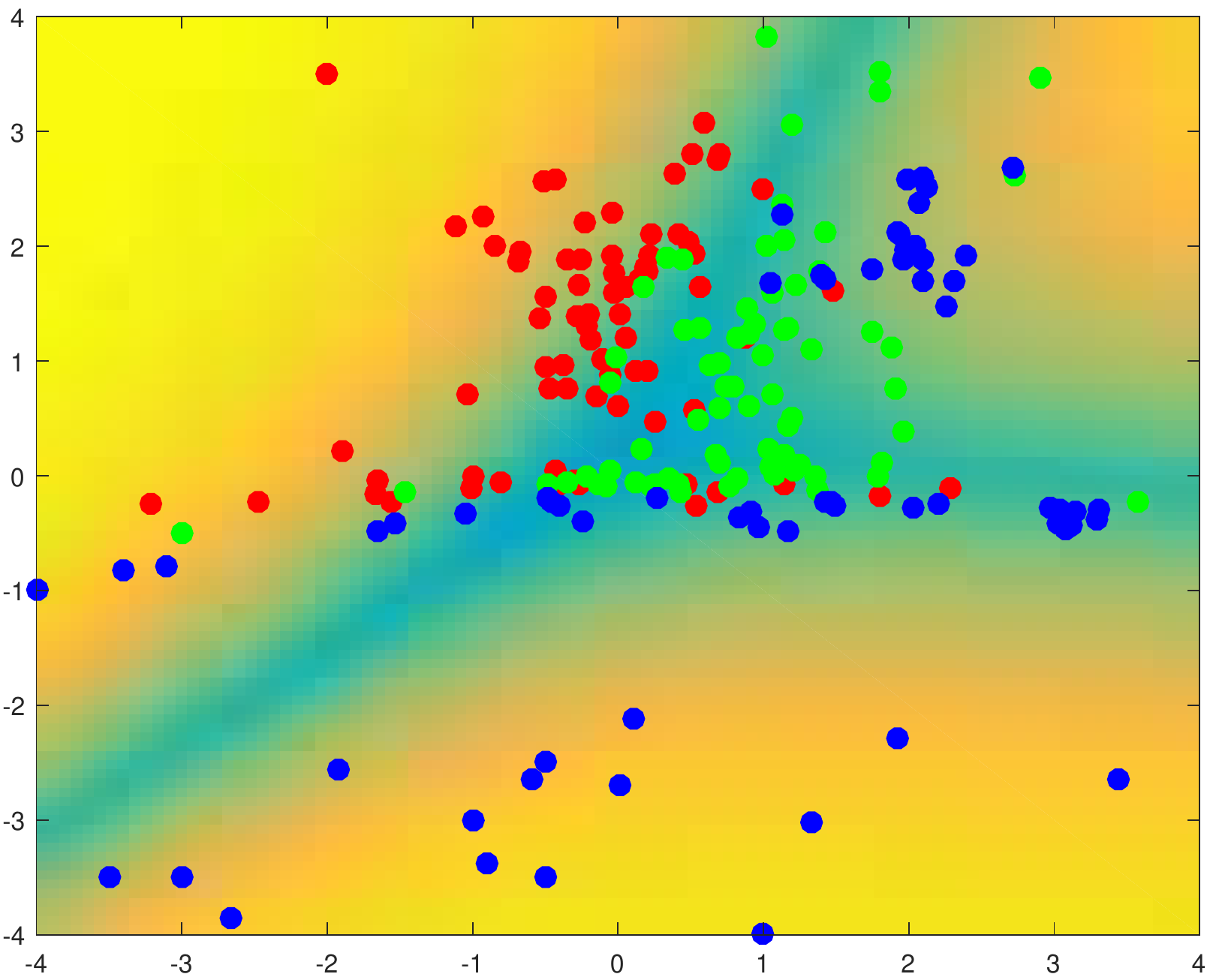}}
  \centerline{(c) Final masked samples}\medskip
\end{minipage}
\caption{(a) Original training samples, (b) Initially masked training samples,
  (c) Final masked samples using Algorithm \ref{a:DFG}}
\label{fig:toy}
\end{figure*}

\begin{figure*}
\begin{minipage}[b]{0.3\linewidth}
  \centering
  \centerline{\includegraphics[width=4.5cm]{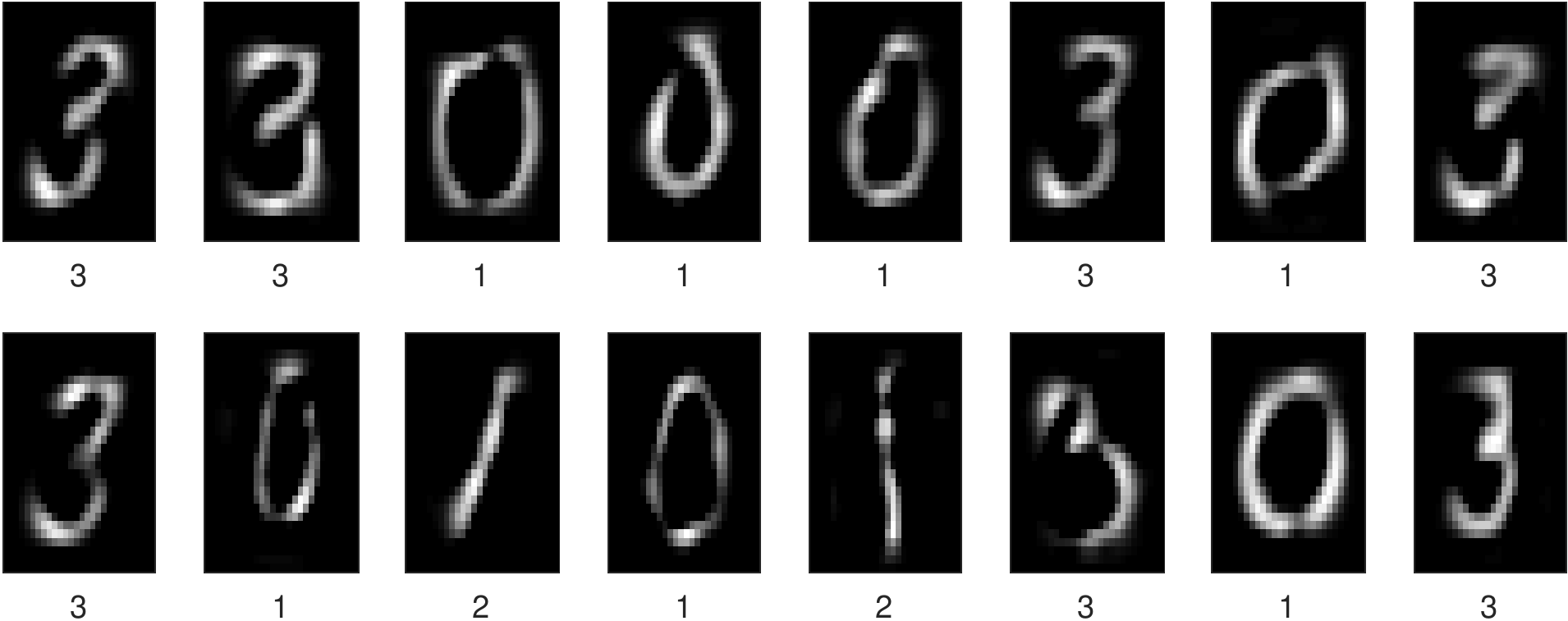}}
  \centerline{(a) True training samples}\medskip
\end{minipage}
\hfill
\begin{minipage}[b]{0.3\linewidth}
  \centering
  \centerline{\includegraphics[width=4.5cm]{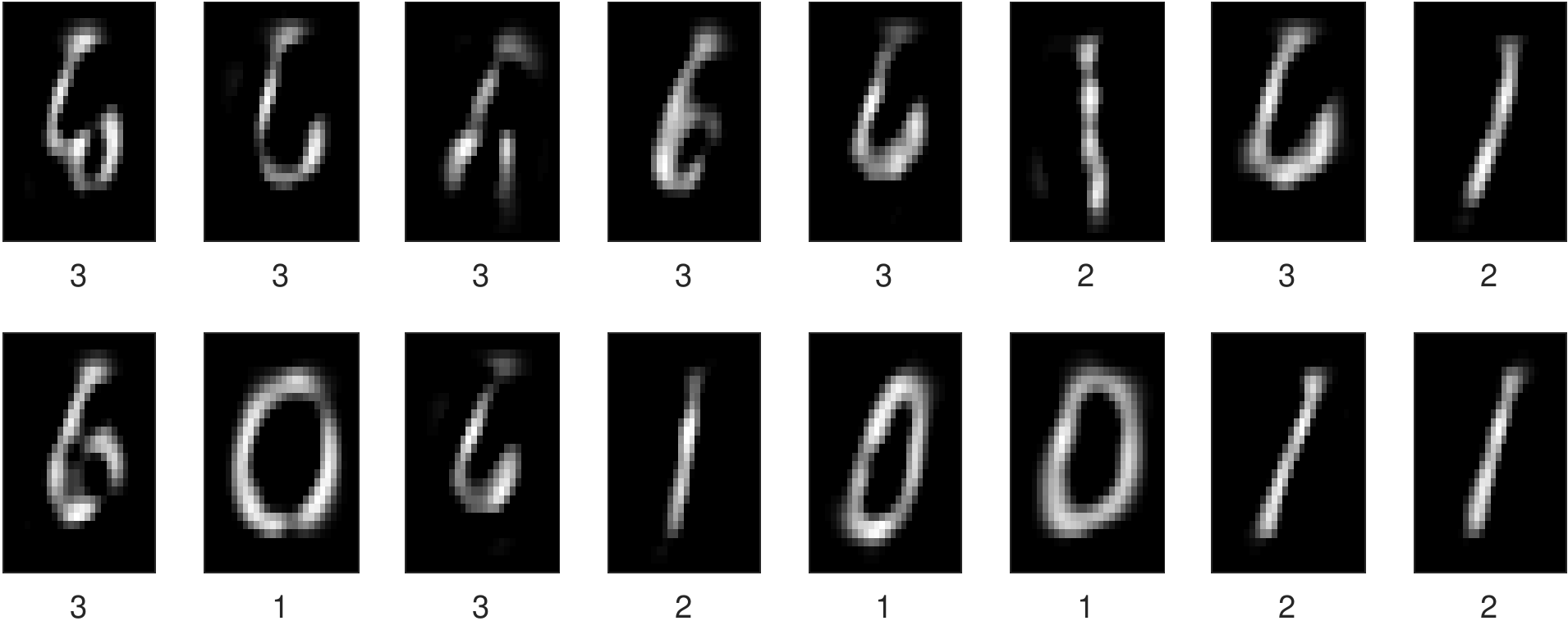}}
  \centerline{(b) Initially masked samples set $\mathbb{S}$}\medskip
\end{minipage}
\hfill
\begin{minipage}[b]{0.3\linewidth}
  \centering
  \centerline{\includegraphics[width=4.5cm]{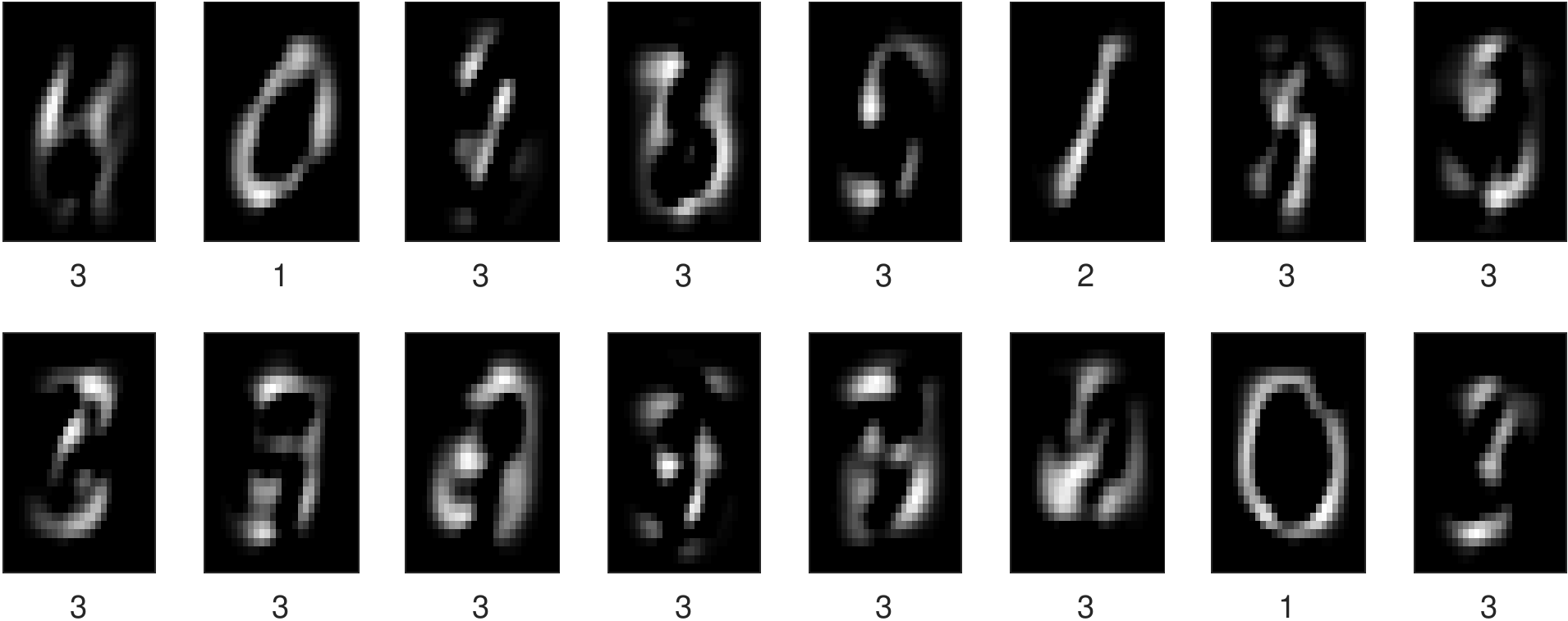}}
  \centerline{(c) Final masked samples using Algorithm \ref{a:DFG}}\medskip
\end{minipage}
\begin{minipage}[b]{0.3\linewidth}
  \centering
  \centerline{\includegraphics[width=4.5cm]{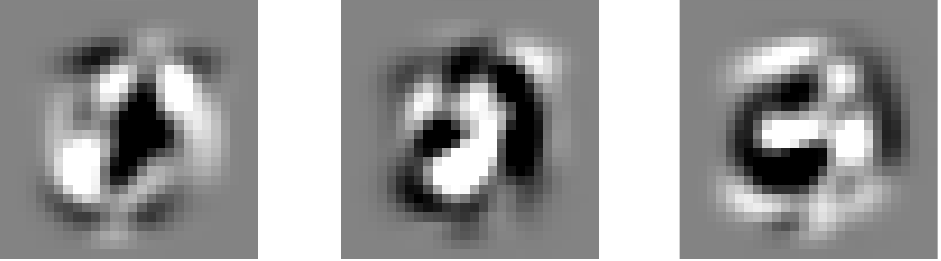}}
  \centerline{(d) True training ${\bf w}$}\medskip
\end{minipage}
\hfill
\begin{minipage}[b]{0.3\linewidth}
  \centering
  \centerline{\includegraphics[width=4.5cm]{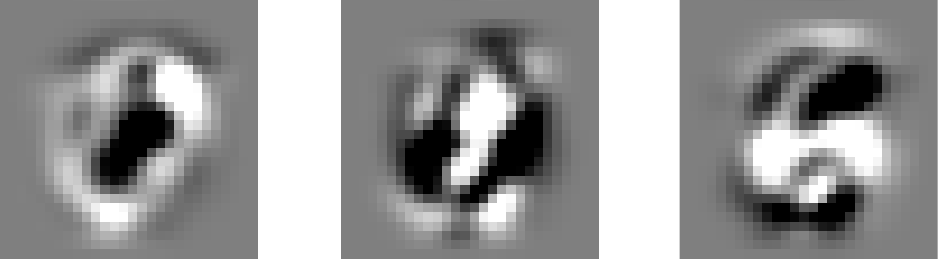}}
  \centerline{(e) Initially masked ${\bf w}$ set $\mathbb{S}$}\medskip
\end{minipage}
\hfill
\begin{minipage}[b]{0.3\linewidth}
  \centering
  \centerline{\includegraphics[width=4.5cm]{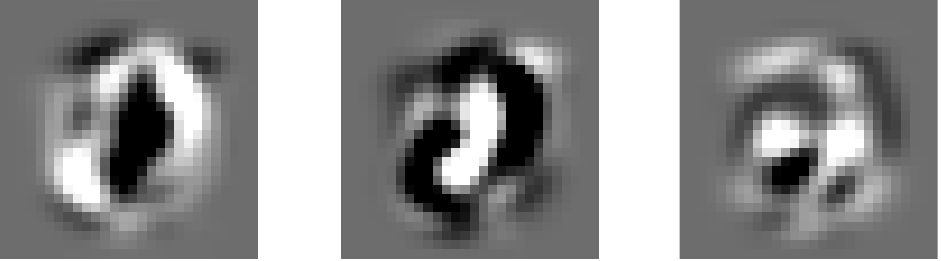}}
  \centerline{(f) Masked ${\bf w}$}\medskip
\end{minipage}
\caption{(a) True training samples, (b) Initially masked samples in $\mathbb{S}$, (c) Final masked samples using Algorithm \ref{a:DFG}, and their corresponding ${\bf w}$'s visualization (d-f) on MNIST
  datasets. We have digits from $0,1,3$ and we would like to replace
  $3$ with $6$ using some fake samples.}
\label{fig:toy1}
\end{figure*}

\begin{figure*}
\begin{minipage}[b]{0.3\linewidth}
  \centering
  \centerline{\includegraphics[width=4.5cm]{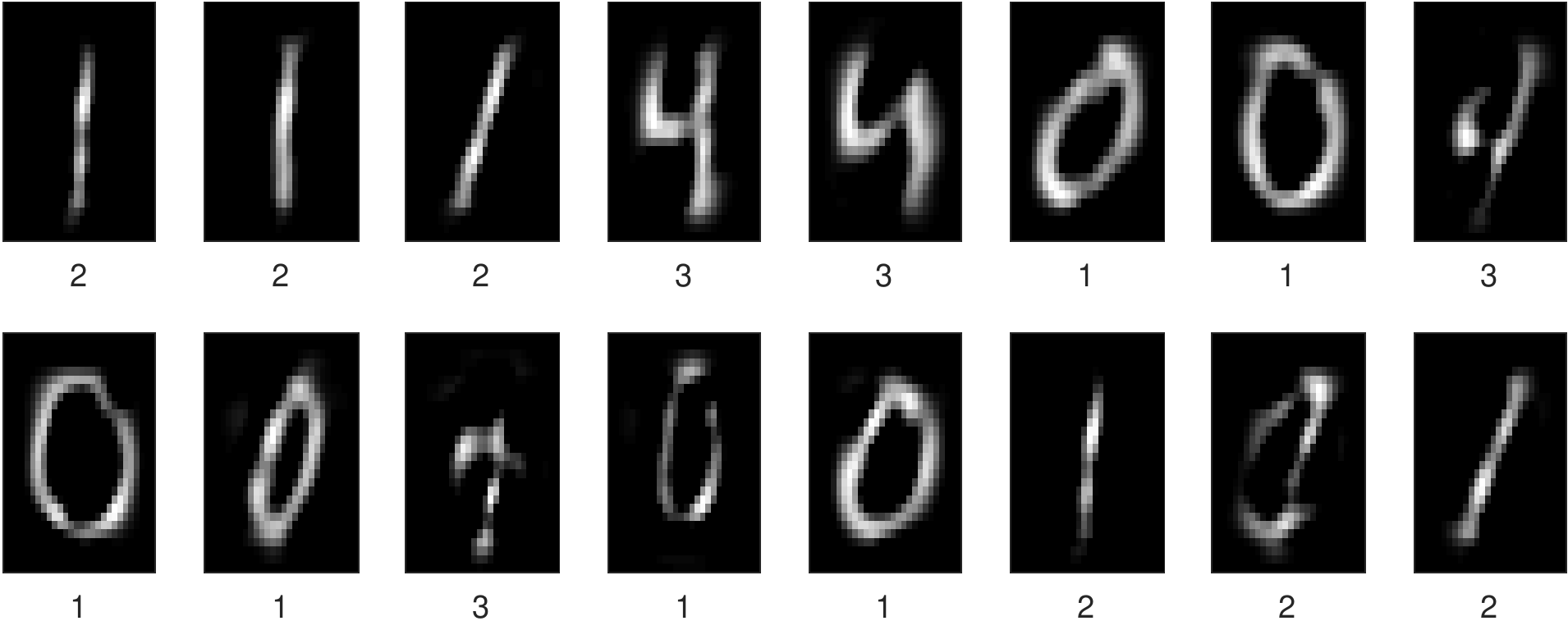}}
  \centerline{(a) True training samples}\medskip
\end{minipage}
\hfill
\begin{minipage}[b]{0.3\linewidth}
  \centering
  \centerline{\includegraphics[width=4.5cm]{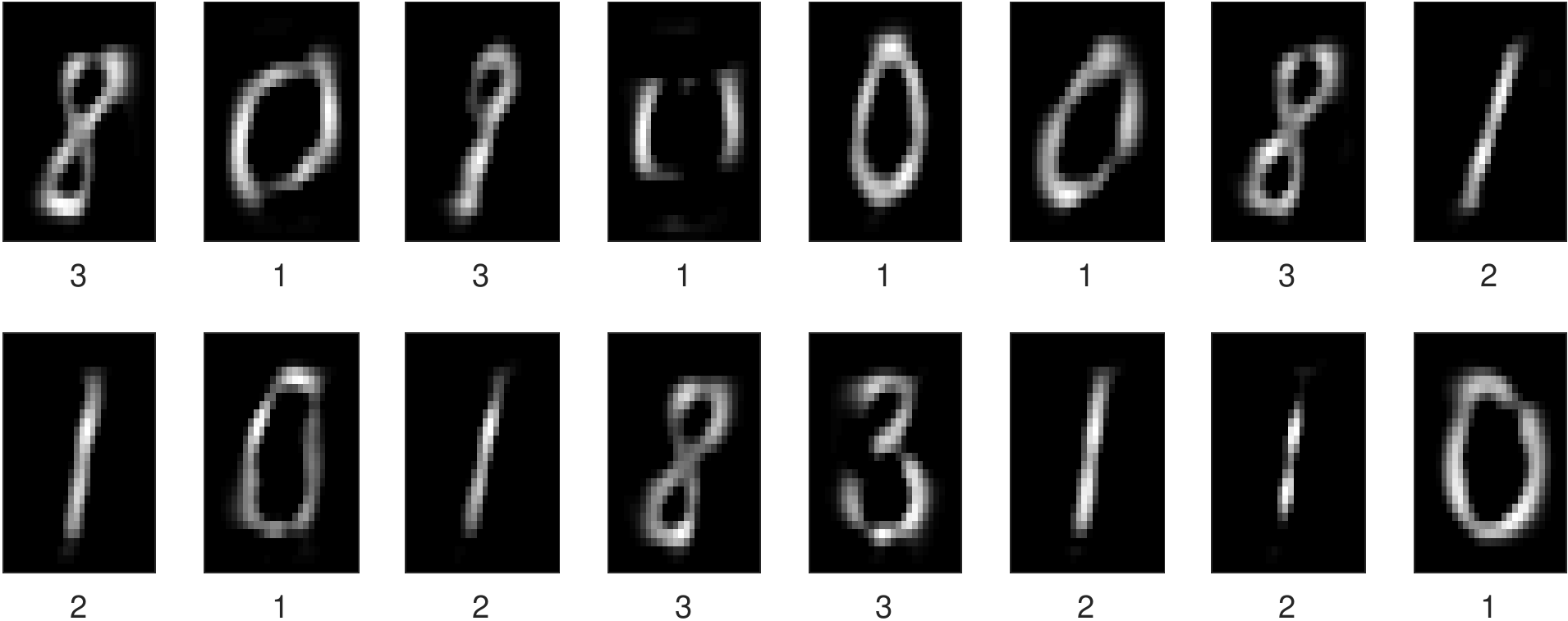}}
  \centerline{(b) Initially masked samples set $\mathbb{S}$}\medskip
\end{minipage}
\hfill
\begin{minipage}[b]{0.3\linewidth}
  \centering
  \centerline{\includegraphics[width=4.5cm]{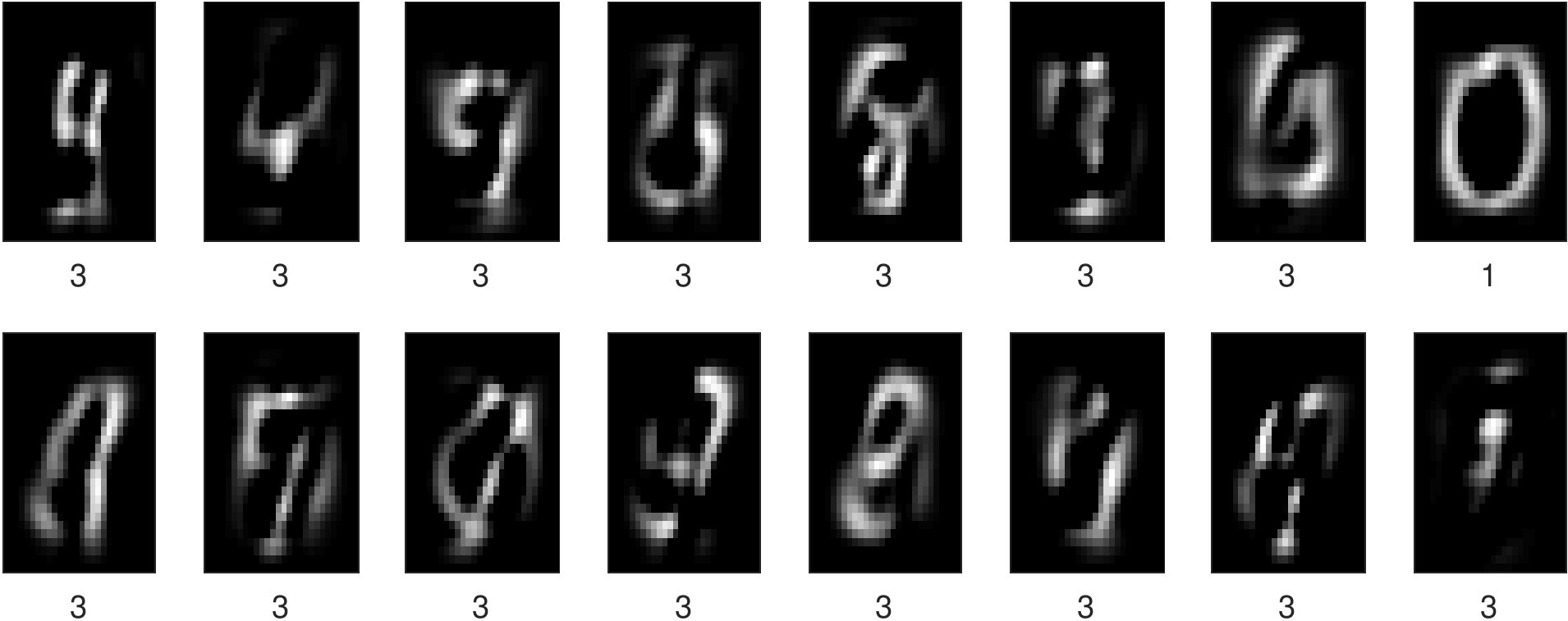}}
  \centerline{(c)  Final masked samples using Algorithm \ref{a:DFG}}\medskip
\end{minipage}
\begin{minipage}[b]{0.3\linewidth}
  \centering
  \centerline{\includegraphics[width=4.5cm]{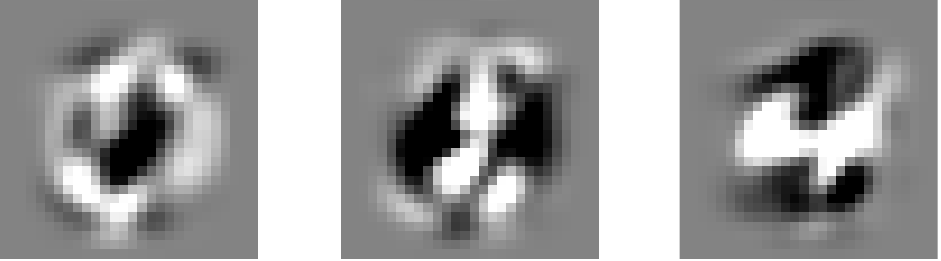}}
  \centerline{(d) True training ${\bf w}$}\medskip
\end{minipage}
\hfill
\begin{minipage}[b]{0.3\linewidth}
  \centering
  \centerline{\includegraphics[width=4.5cm]{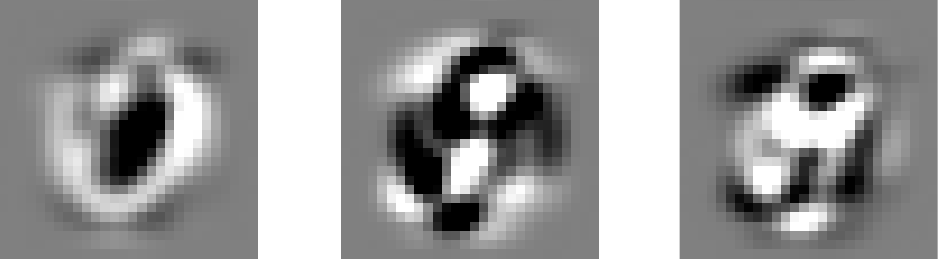}}
  \centerline{(e) Initially masked ${\bf w}$}\medskip
\end{minipage}
\hfill
\begin{minipage}[b]{0.3\linewidth}
  \centering
  \centerline{\includegraphics[width=4.5cm]{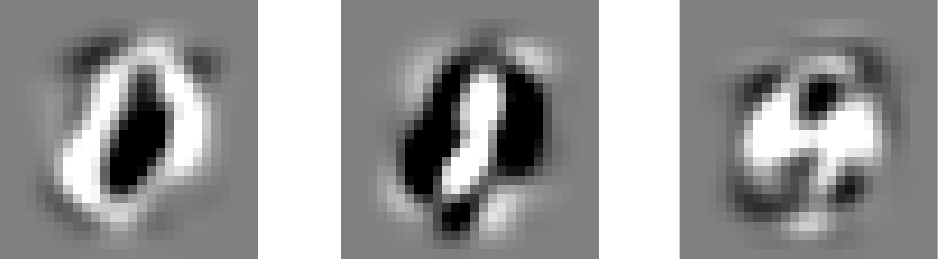}}
  \centerline{(f) Masked ${\bf w}$}\medskip
\end{minipage}
\caption{(a) True training samples, (b) Initially masked samples in $\mathbb{S}$,
and (c) Final masked samples using Algorithm \ref{a:DFG}, and their corresponding ${\bf w}$'s visualization (d-f) on MNIST
  datasets. We have digits from $0,1,4$ and we would like to replace $4$ with $8$ using some fake samples.}
\label{fig:toy2}
\end{figure*}

\subsection{Results on toy data}

\noindent {\bf Datasets:} In this section, the effectiveness of the
proposed method is illustrated on a 2D toy dataset. We sample
$100$ training samples from three normal distributions. The
$1$st class comes from $\mathcal{N}([0; 1.5]^T,0.25I)$, the $2$nd
class comes from $\mathcal{N}([1; 1]^T,0.25I)$, and the $3$rd class
comes from $\mathcal{N}([1;-1]^T,0.25I)$, as shown in
Fig.~\ref{fig:toy}(a). Assume that samples from the $3$rd class is sensitive.

\noindent {\bf Setting:}
We initialize the samples in the masked dataset from a
class with a different manifold for the $3$rd class. In particular, we first add to the
published dataset a fake class $3$ with a totally different
distribution manifold from the original class $3$, e.g.,
$\mathcal{N}([2;2]^T,0.25I)$ instead of $\mathcal{N}([1;-1]^T,0.25I)$,
as shown in Fig.~\ref{fig:toy}(b). We then run the masked data
generation method with non-empty training samples set $\mathbb{S}$ as
in Algorithm \ref{a:DFG}.

\noindent {\bf Results:} The samples generated from the proposed
method are shown in Fig.~\ref{fig:toy}(c). From Fig.~\ref{fig:toy}(c),
to accommodate for the shift in distribution manifold of class 3 from
$[2; 2]$ to $[1;-1]$, many other fake samples of class $3$ are added
in the bottom of Fig.~\ref{fig:toy}(c).From Fig.~\ref{fig:toy}(c), we observe the usefulness of regularization,
since less masked samples are on the boundary. From
Fig.~\ref{fig:toy}(c) and Fig.~\ref{fig:toy}(a), the generated
samples from class $3$ is significantly different from the original
true samples from class $3$, which implies that the data is private.
However, the resulting classifier or the boundary learned from the
three classes are almost similar for original data and published data. As a result, users are still able to access original real data from classes $1$ and $2$, and at the same time achieve the classifier for class $3$ which is private now.

\subsection{Results on MNIST digits data}
In this section, we consider the effectiveness of the proposed
algorithm on the MNIST handwritten digit dataset.

\noindent {\bf Datasets:} We use PCA to reduce the dimensionality of
the data to 25. Similar to the toy example, we select samples from three
digits, e.g., three digits $\{0, 1, 3\}$ as in Fig.~\ref{fig:toy1}(a),
and three digits $\{0, 1, 4\}$ as in Fig.~\ref{fig:toy2}(a). The
corresponding classifier learned from three digits $\{0, 1, 3\}$ is
shown in Fig.~\ref{fig:toy1}(d), and from three digits $\{0,1,4\}$ is
shown in Fig.~\ref{fig:toy2}(d). \footnote{For visualization of a classifier, e.g., in  Fig.~\ref{fig:toy1}(d-f), we project the classifier of each class back to the two dimensional space.} From those figures, e.g., in
Fig.~\ref{fig:toy1}(d), the visualized classifier represents the three
corresponding digits $\{0, 1, 3\}$.

\noindent {\bf Setting:} We first explain how to generate a non-empty
initially masked training samples $\mathbb{S}$ in Algorithm \ref{a:DFG}. In
particular, the first two digits from the initially masked training samples
are the same as the two digits of the original training samples. For
example, we still uses samples from digits $0$ and $1$ for initially masked
training samples as in Fig.~\ref{fig:toy1}(b). However, for the last
digit of the initially masked training samples, we use a totally
different digit from the last digit of the true training samples. For example, we use digit $6$ instead of digit $3$ as the last digit as in
Fig.~\ref{fig:toy1}(b). The corresponding classifier learned from the
initially masked training samples $\mathbb{S}$ is visualized in
Fig.~\ref{fig:toy2}(e). 

\noindent {\bf Result:} We then iteratively add masked training
samples into $\mathbb{S}$ using the masked data generation method in Algorithm \ref{a:DFG}. The masked
samples generated by Algorithm \ref{a:DFG} into $\mathbb{S}$ are shown
in Fig.~\ref{fig:toy1}(c). Note that several samples among them remove
the effect of digit $6$, e.g., the $4$th sample from the left in the
first row of Fig.~\ref{fig:toy1}(c). On another hand, several among
them add the effect of digit $3$ back to the classifier, e.g., the
image at the bottom right of Fig.~\ref{fig:toy1}(c). Moreover, because
of the adding masked samples, the classifier learned from the masked training samples
is similar to the original classifier learned from the original
training samples. For example, the classifier in
Fig.~\ref{fig:toy1}(f) is similar to the classifier in
Fig.~\ref{fig:toy1}(d).

A similar visualization example is shown in Fig.~\ref{fig:toy2}, where
the original training samples are  digit $\{0, 1, 4\}$ as in
Fig.~\ref{fig:toy2}(a), the initially masked training samples in
$\mathbb{S}$ are  digit $\{0, 1, 8\}$ as in
Fig.~\ref{fig:toy2}(b), and after generating masked samples, the
classifier of the masked data as in Fig.~\ref{fig:toy2}(f) is similar
to the classifier of original data as in Fig.~\ref{fig:toy2}(d).

\begin{figure*}[tp]
\begin{minipage}[b]{0.33\linewidth}
  \centering
  \centerline{\includegraphics[width=5.5cm]{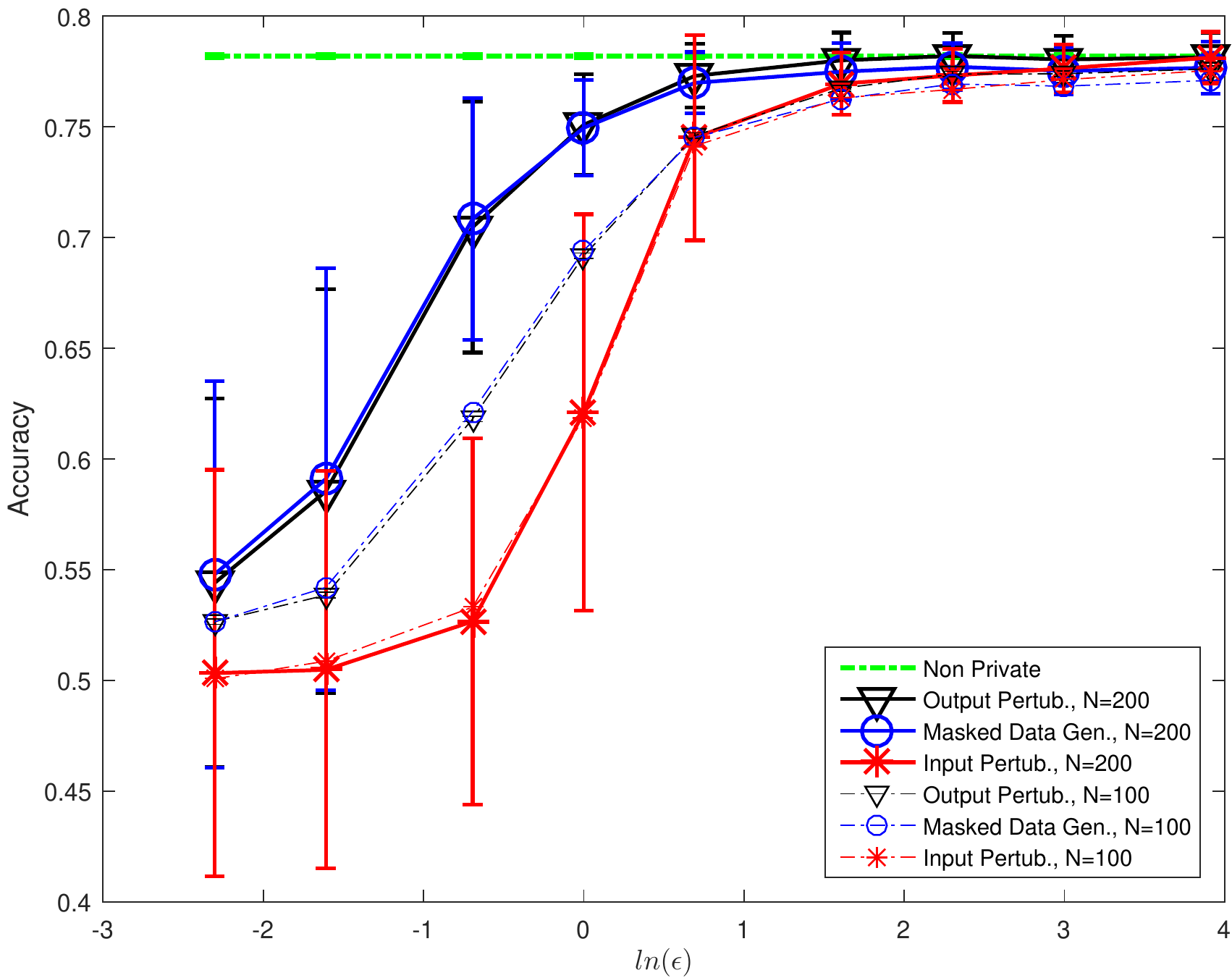}}
  \centerline{(a) Adult income}\medskip
\end{minipage}
\hfill
\begin{minipage}[b]{0.33\linewidth}
  \centering
  \centerline{\includegraphics[width=5.5cm]{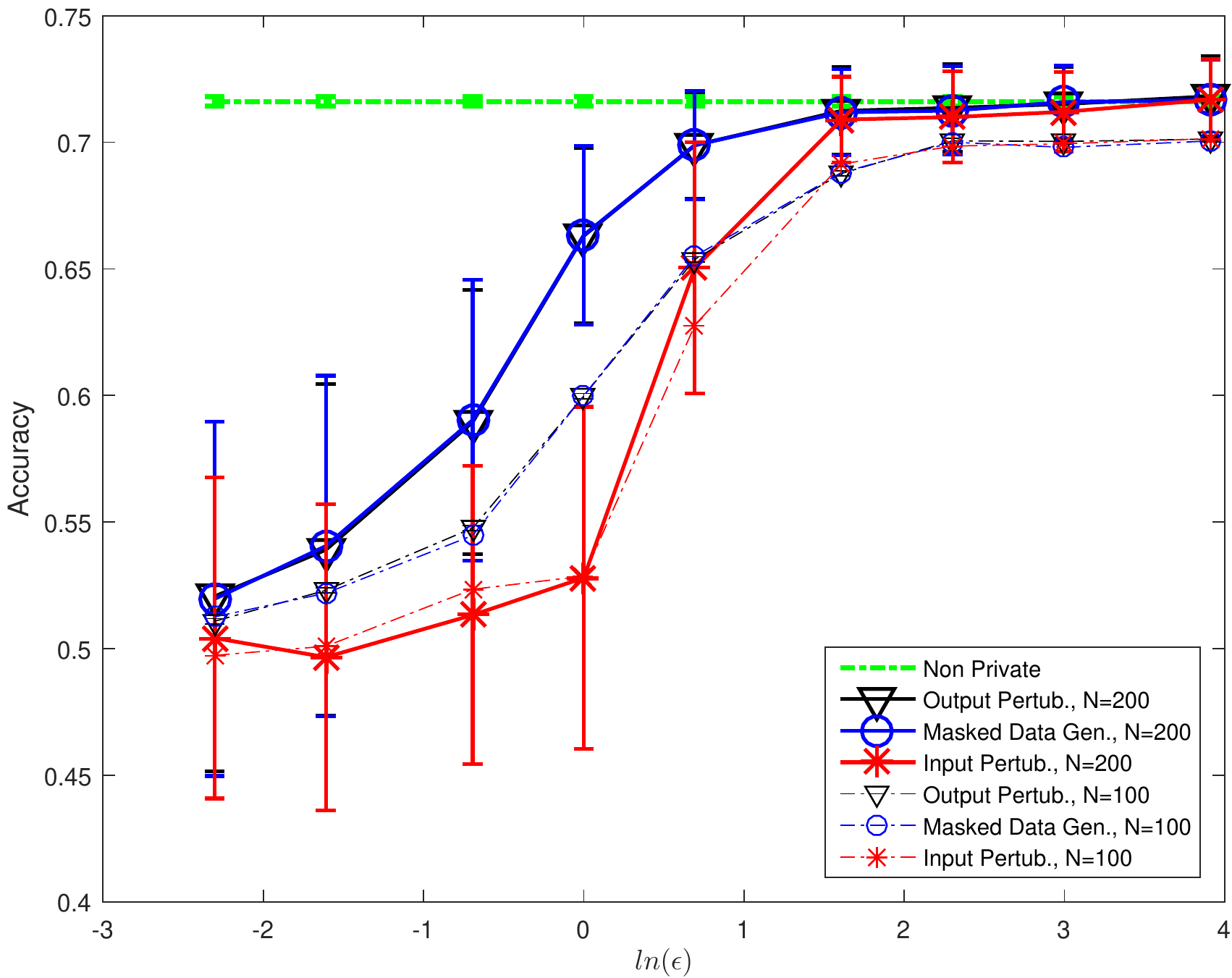}}
  \centerline{(b) German credit}\medskip
\end{minipage}
\hfill
\begin{minipage}[b]{0.33\linewidth}
  \centering
  \centerline{\includegraphics[width=5.5cm]{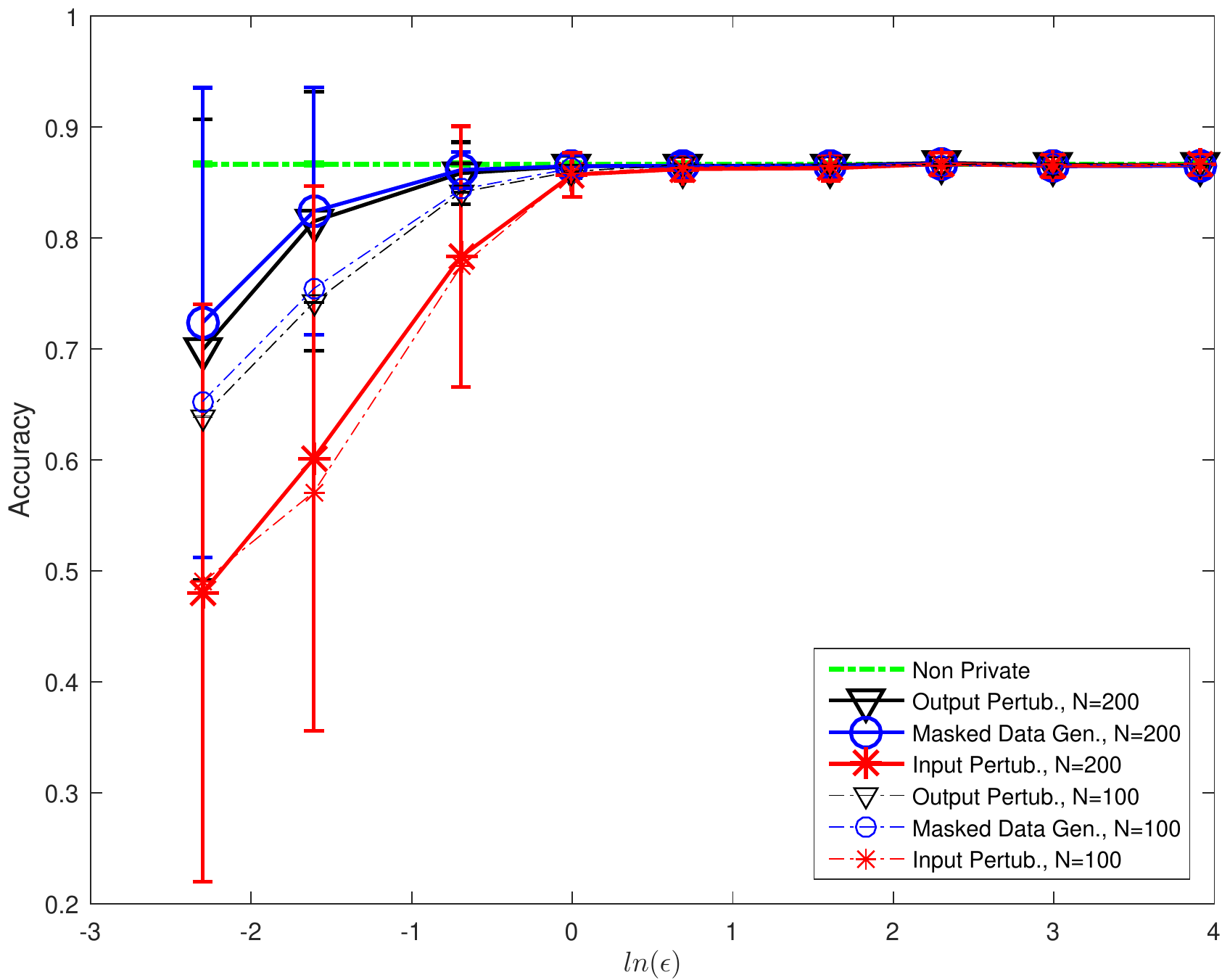}}
  \centerline{(c) Age of Abalone}\medskip
\end{minipage}
\begin{minipage}[b]{0.33\linewidth}
  \centering
  \centerline{\includegraphics[width=5.5cm]{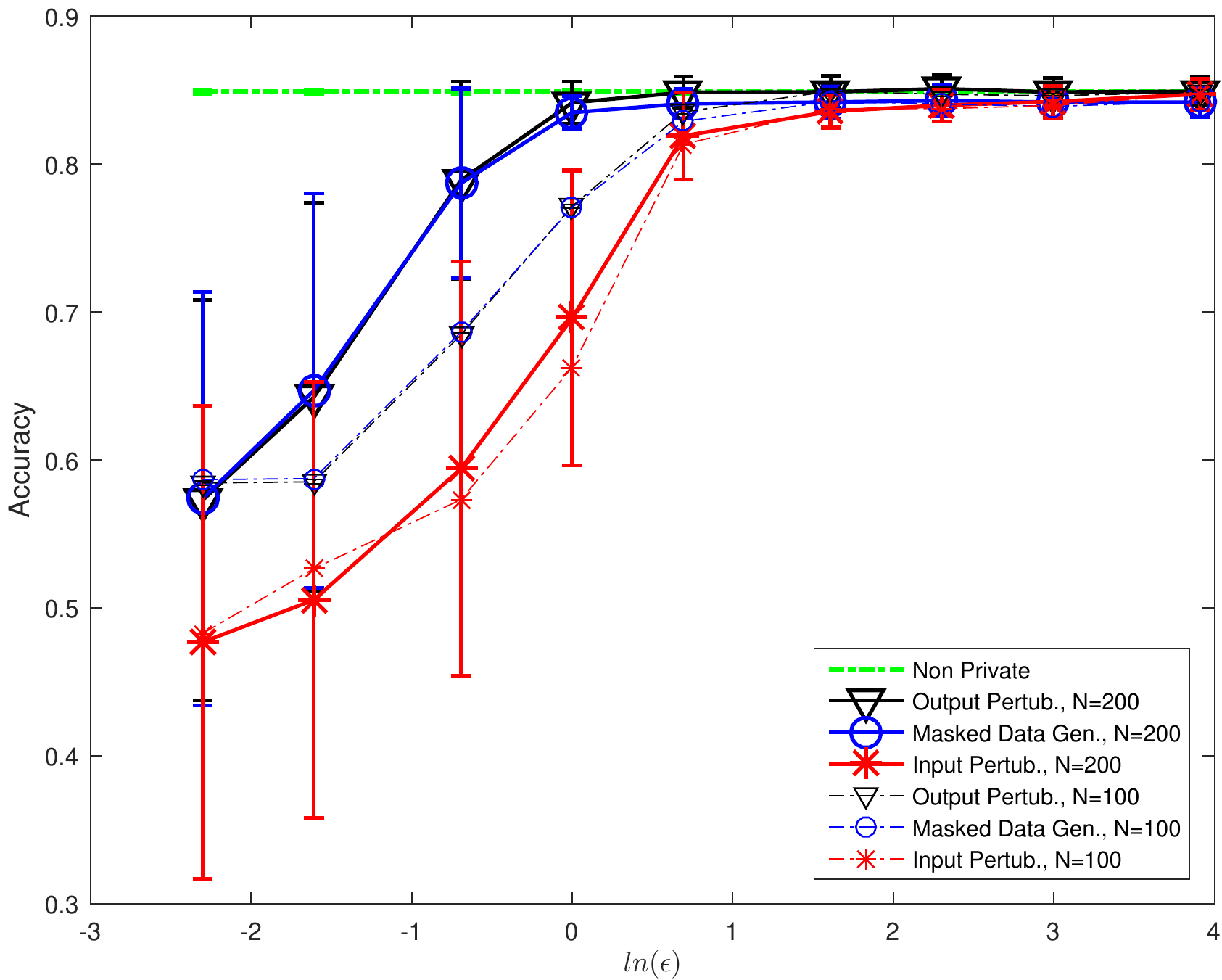}}
  \centerline{(d) Wave form}\medskip
\end{minipage}
\hfill
\begin{minipage}[b]{0.33\linewidth}
  \centering
  \centerline{\includegraphics[width=5.5cm]{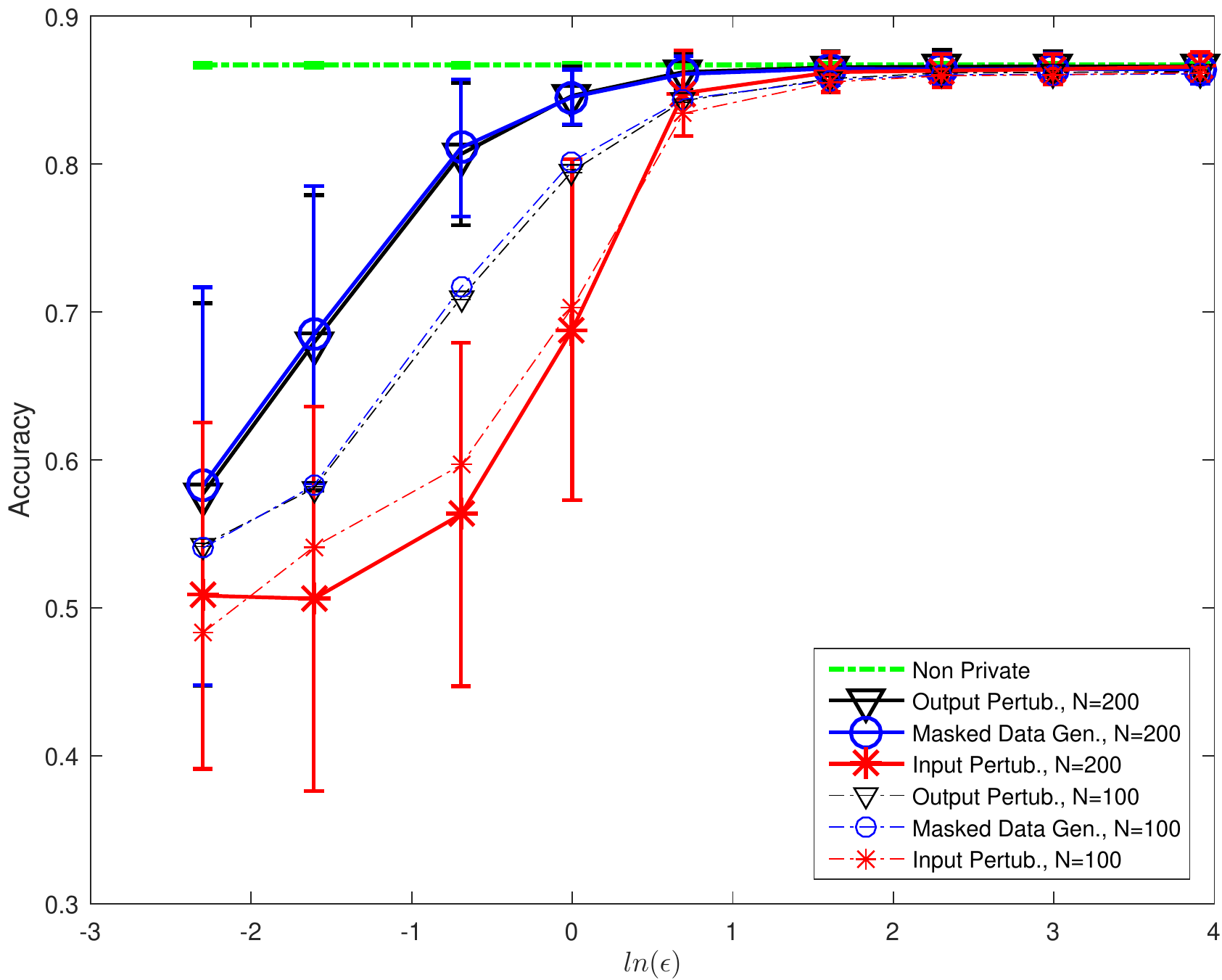}}
  \centerline{(e) AUS credit}\medskip
\end{minipage}
\hfill
\begin{minipage}[b]{0.33\linewidth}
  \centering
  \centerline{\includegraphics[width=5.5cm]{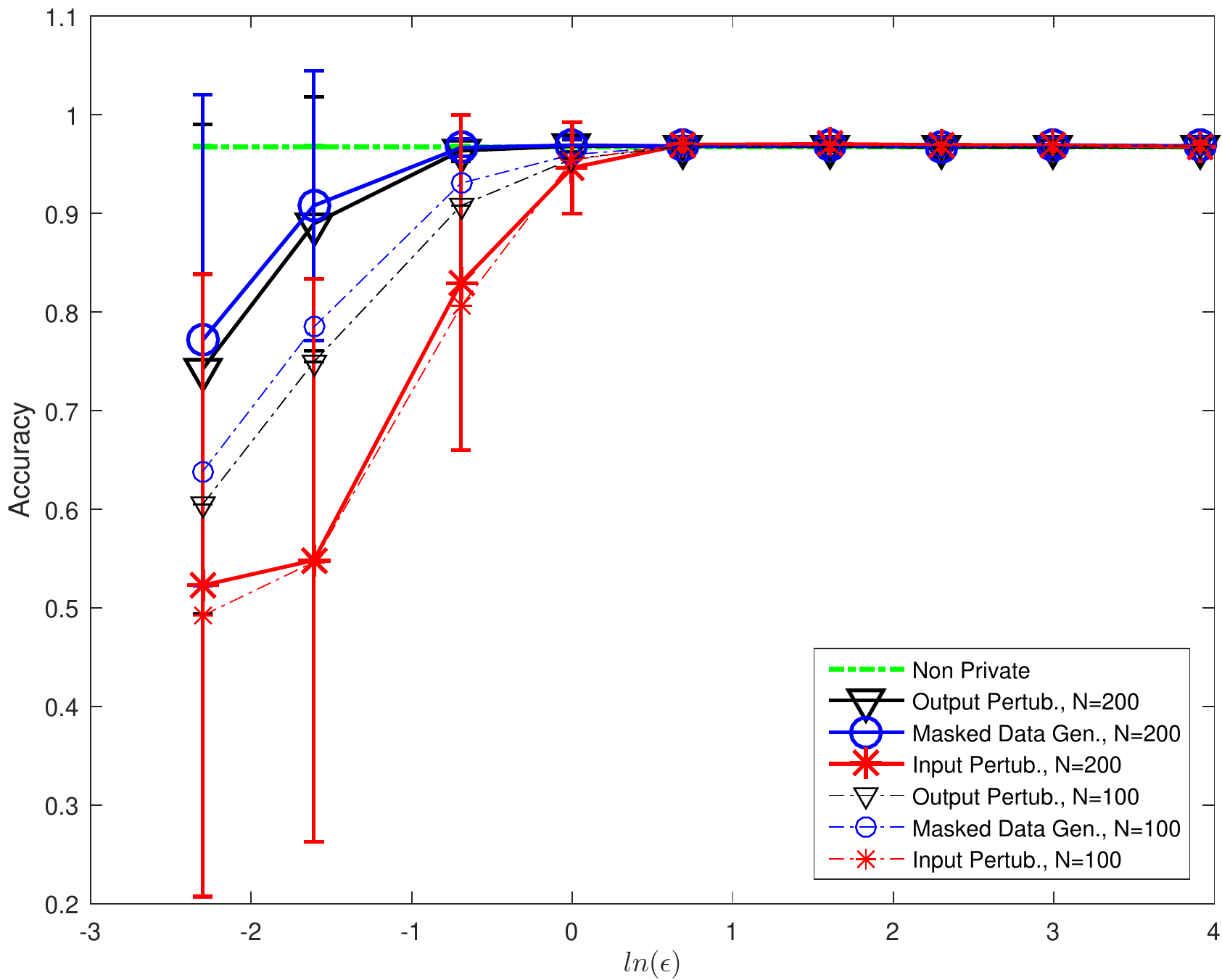}}
  \centerline{(f) Breast cancer}\medskip
\end{minipage}
\begin{minipage}[b]{0.33\linewidth}
  \centering
  \centerline{\includegraphics[width=5.5cm]{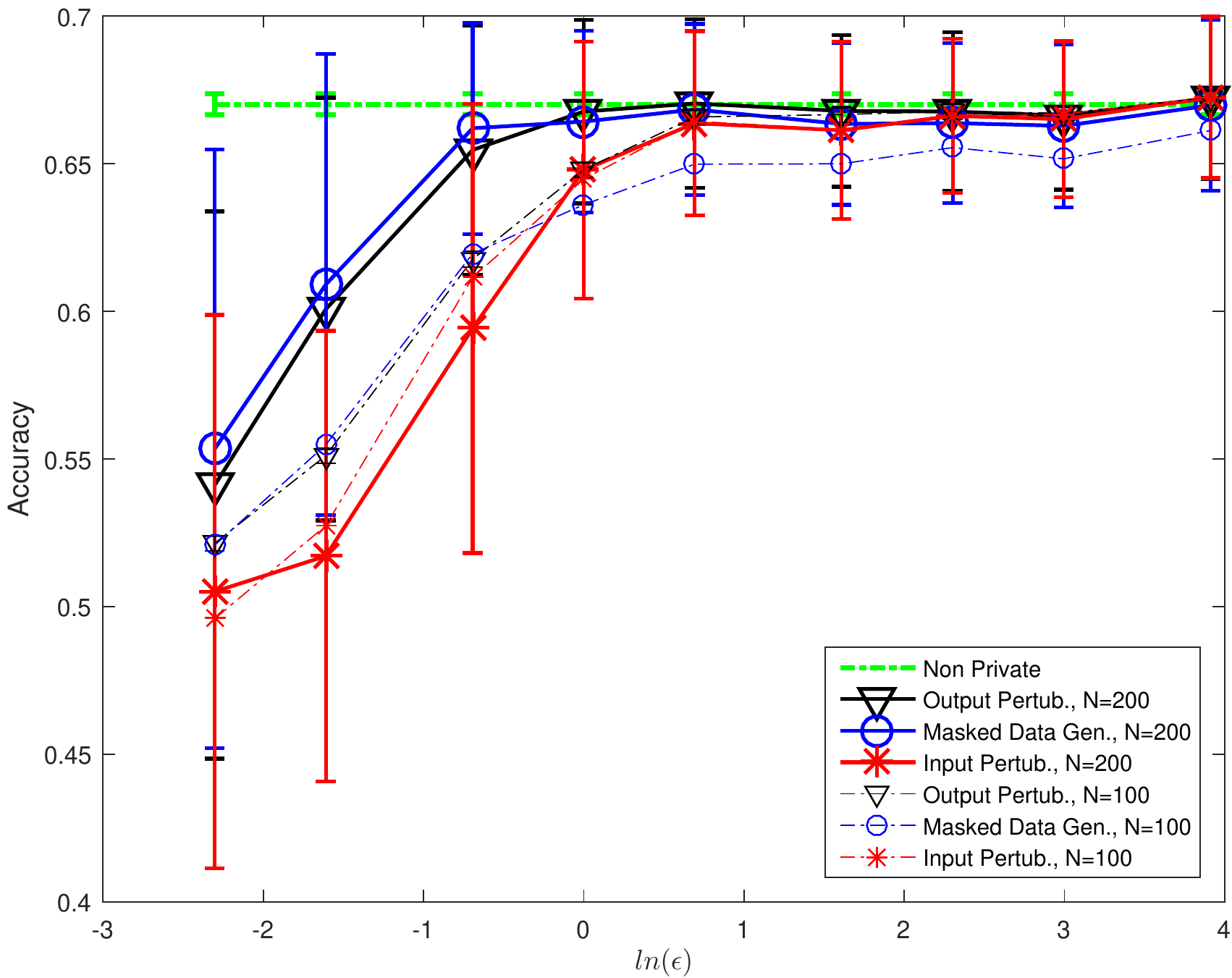}}
  \centerline{(g) Blood transfusion}\medskip
\end{minipage}
\begin{minipage}[b]{0.33\linewidth}
  \centering
  \centerline{\includegraphics[width=5.5cm]{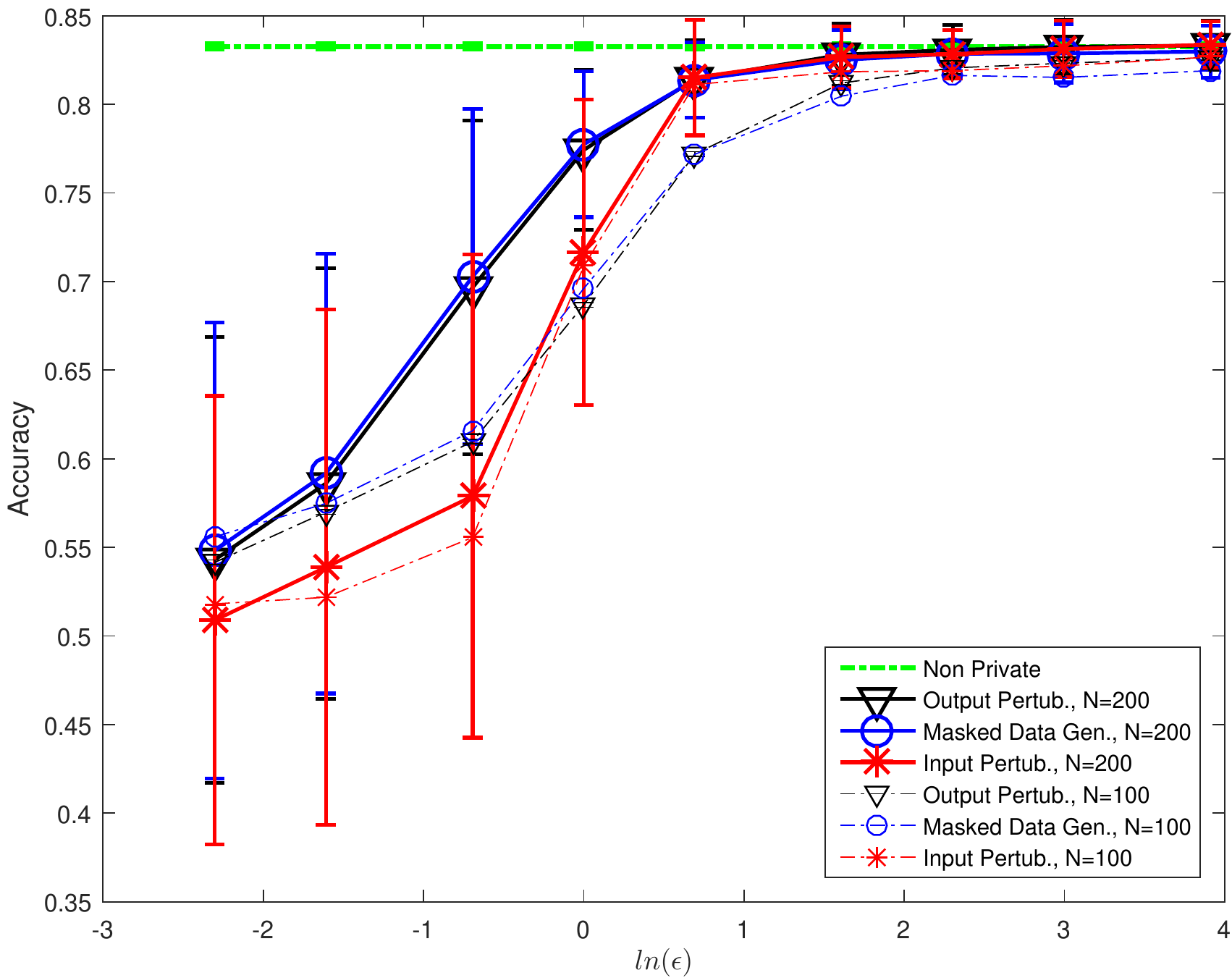}}
  \centerline{(h) Heart disease}\medskip
\end{minipage}
\hfill
\begin{minipage}[b]{0.33\linewidth}
  \centering
  \centerline{\includegraphics[width=5.5cm]{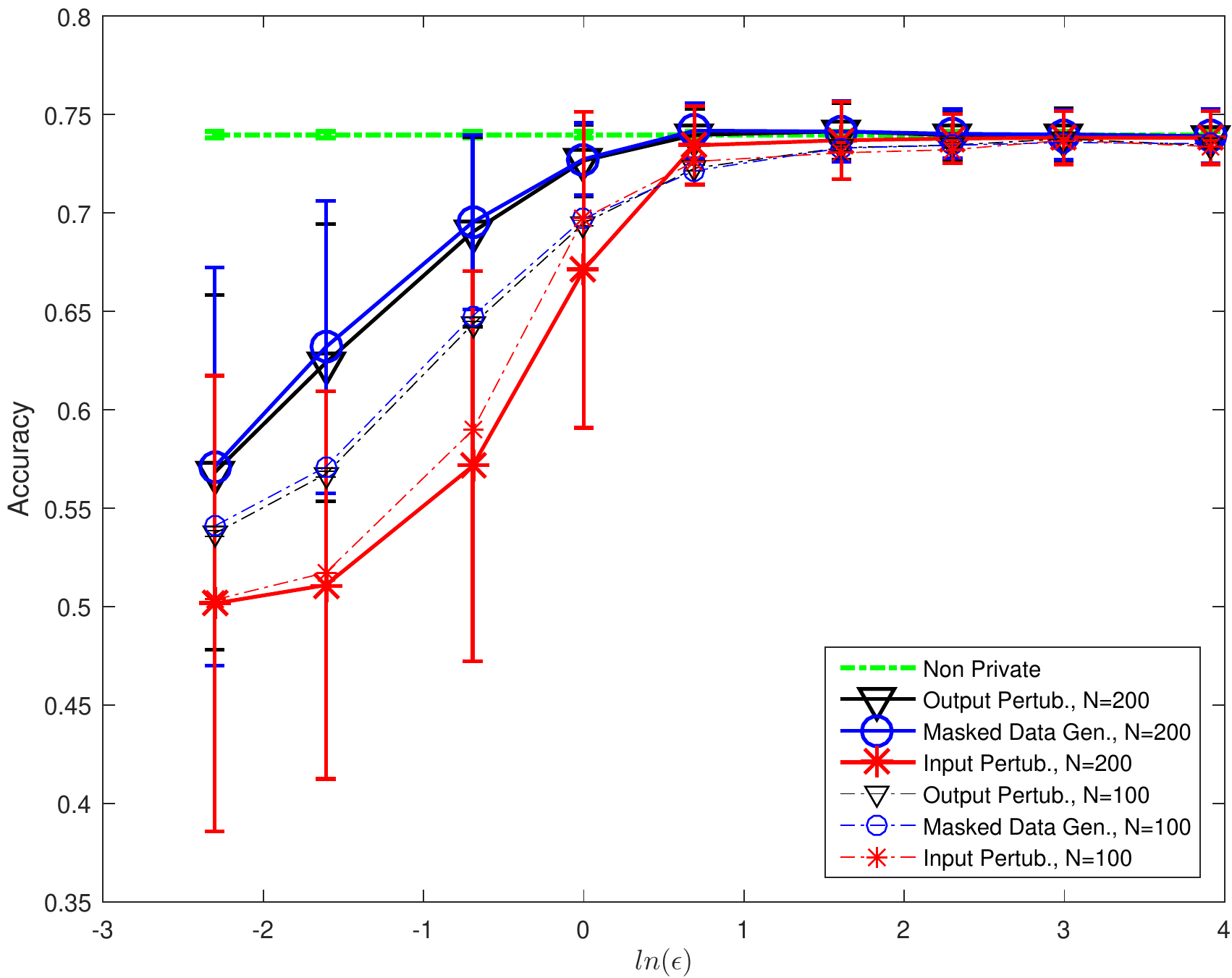}}
  \centerline{(k) Diabetics }\medskip
\end{minipage}
\begin{minipage}[b]{0.33\linewidth}
  \centering
  \centerline{\includegraphics[width=5.5cm]{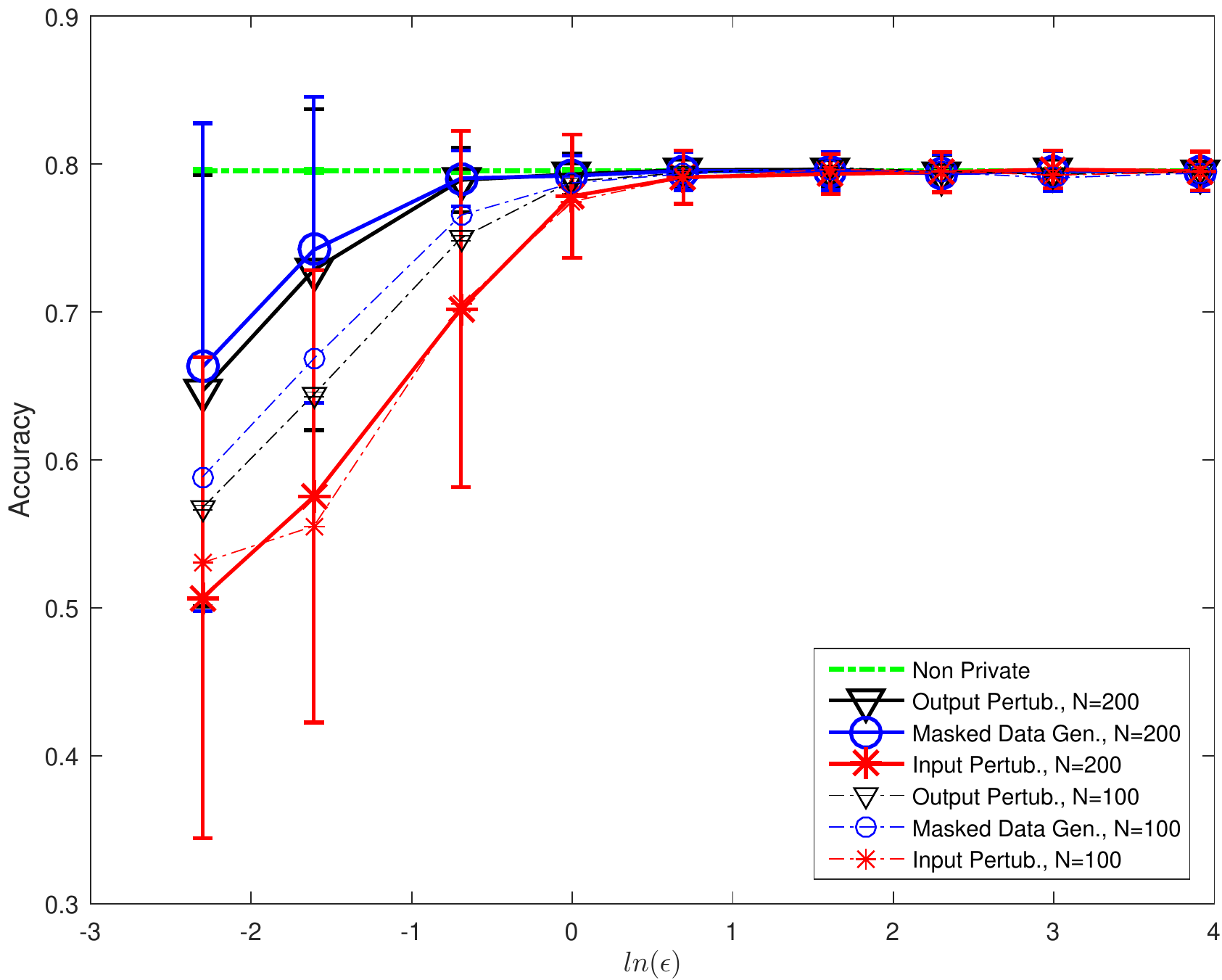}}
  \centerline{(i) Mammographic }\medskip
\end{minipage}
\hfill
\begin{minipage}[b]{0.33\linewidth}
  \centering
  \centerline{\includegraphics[width=5.5cm]{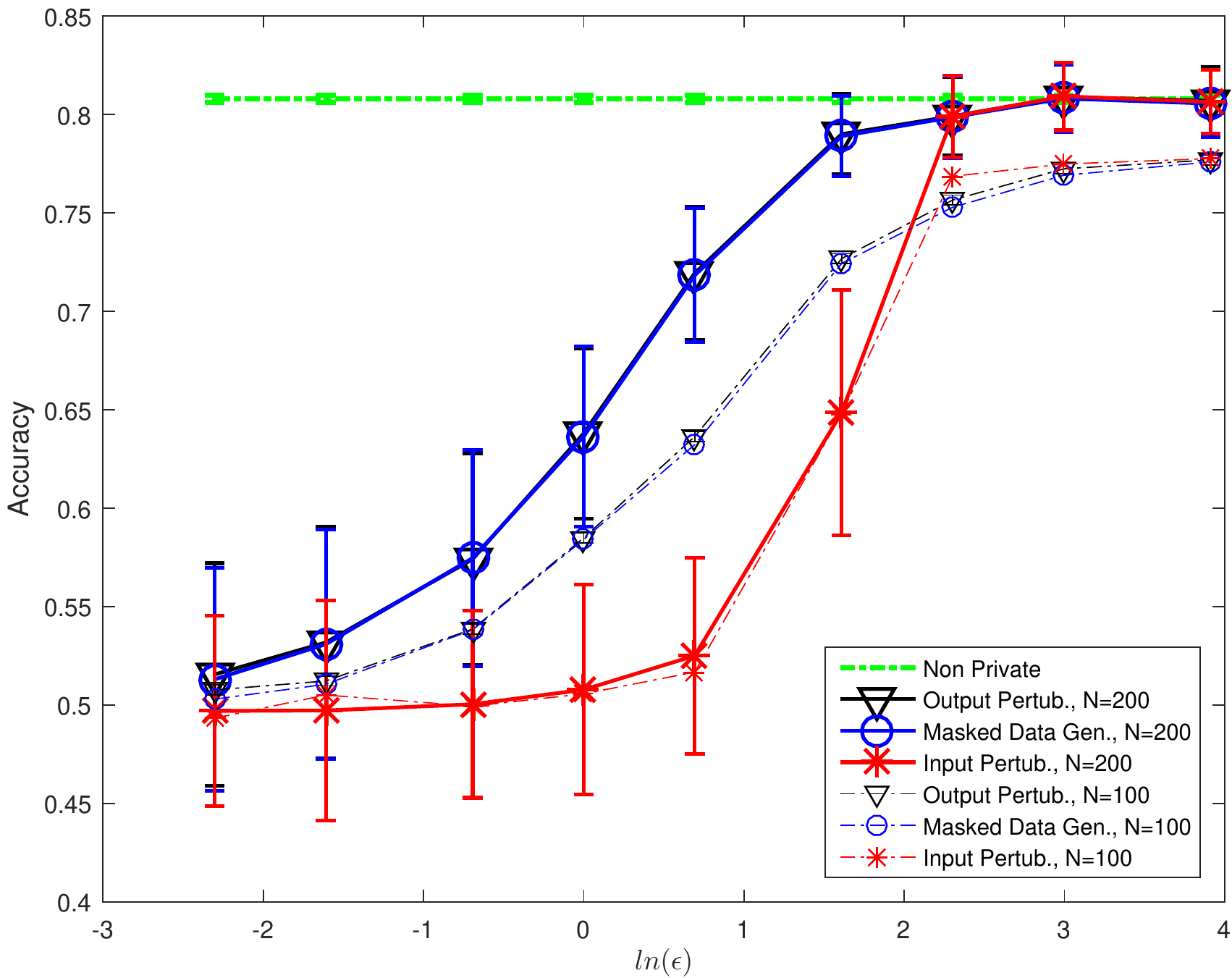}}
  \centerline{(j) SpliceDNA }\medskip
\end{minipage}
\hfill
\begin{minipage}[b]{0.33\linewidth}
  \centering
  \centerline{\includegraphics[width=5.5cm]{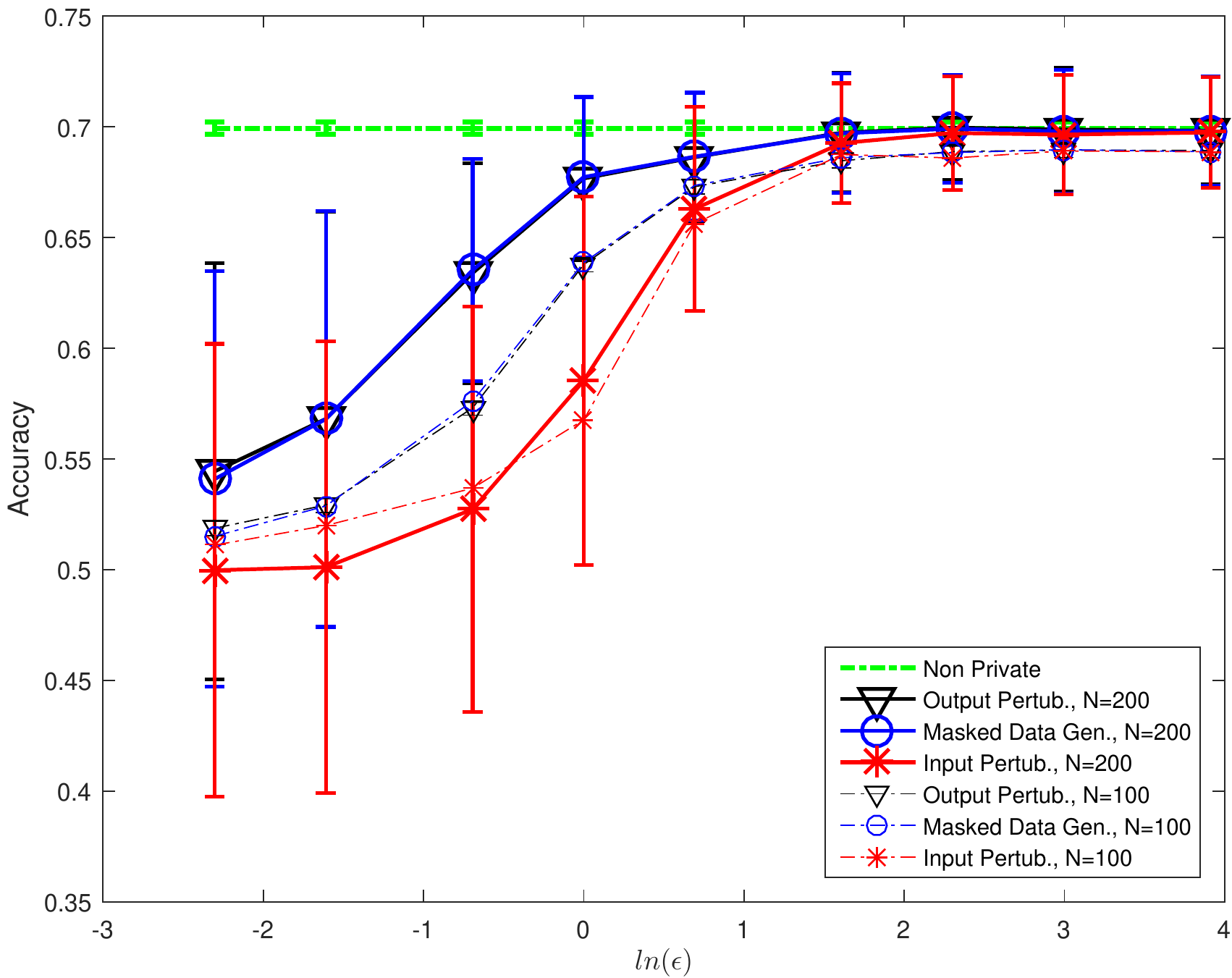}}
  \centerline{(m) Image }\medskip
\end{minipage}

\caption{The accuracy and privacy ($\epsilon$ in log-scale) trade off for $12$ benchmark datasets.}
\label{fig:Real}
\end{figure*}

\subsection{Results on UCI datasets}

\noindent {\bf Datasets:} We demonstrate the effectiveness of the proposed method on several UCI datasets in sensitive domains.

\noindent {\bf Evaluation Measure:} For all datasets, we uniformly select a validation set $V=\{{\bf
  x}_i\}_{i=1}^{n_{val}}$ of samples from two classes. We denote the
ground truth labels for these samples as $L_{true}$. Using ${\bf
  w}'$, the classifier trained on masked data, we predict the labels for the validation set, namely $V_{masked}$. Then, we compute the accuracy of ${\bf w}'$ as the fraction of cases
where $V_{masked}$ matches $L_{true}$.

\noindent {\bf Setting:} We consider the regularization parameter
$\lambda=0.5$. Moreover, to evaluate the effectiveness of the proposed method and the input perturbation method when the number of training samples increases, we consider two cases: $N=100$ and $N=200$. We vary the value
of $\epsilon$ in the set $\{0.1, 0.2, 0.5, 1, 2, \dots, 20, 50\}$,
e.g., log-scale. For each value of $\epsilon$, we generate $50$
training datasets, run the proposed masked data generation Algorithm
\ref{a:DFG} and the input perturbation Algorithm \ref{a:IPB} on each dataset, then
report the mean and standard deviation accuracy of both algorithms. We also evaluate the accuracy using the classifier after adding Laplace noise, i.e., after Step 2 of Algorithm \ref{a:DFG}, which is named as output perturbation.

\noindent {\bf Analysis:} As shown in Fig.~\ref{fig:Real}, first, as
$\epsilon$ increases, the accuracy of both methods
increase. Additionally, for a particular value of $\epsilon$, the proposed
method works better than input perturbation algorithm. Moreover, as $N$ increases from $100$ to $200$, the proposed method gets higher accuracy for the same value of $\epsilon$. In contrast, the accuracy of the input perturbation method does not change much as $N$ increases. Furthermore, note that the
input perturbation method only updates the data independently from the
machine learning model. In contrast, the data generated by the proposed method is
directly tied to the model, e.g., logistic regression with a particular value of $\lambda$, which may lead to higher accuracy. Moreover, the performance of the classifier trained on masked samples is comparable to those of the classifier trained on original training samples then adding Laplace noise, i.e., after Step 2 of Algorithm \ref{a:DFG}. The results indicate that the proposed masked data generation Algorithm \ref{a:DFG} is able to create masked samples with corresponding classifier close to the perturbed classifier.

\section{Conclusions}
In this paper, we proposed a data masking technique for
privacy-sensitive learning. The main idea is to iteratively find
masked data such that the gradient of the likelihood on the classifier
with regarding to the masked data is zero. Our theoretical analysis
showed that the proposed technique achieves higher utility compared to a
traditional input perturbation technique.  Experiments on multiple real-world
datasets also demonstrated the effectiveness of the proposed method.

\section{Appendices}

\noindent \textbf{Proof for Proposition \ref{l:4}.} Assume there are two training datasets
$\mathbb{D}_{1}=\{{\bf x}_{1i}\}_{i=1}^N$ and
$\mathbb{D}_{2}=\{{\bf x}_{2i}\}_{i=1}^N$, which are different
at only one sample. Without the loss of generality, we
assume ${\bf x}_{1i}={\bf x}_{2i}$ for $i=\{1,
2,\dots,N-1\}$, and ${\bf x}_{1N}\neq {\bf x}_{2N}$. Assume the outputs of Algorithm \ref{a:DFG} is
$\mathbb{O}=\{{\bf x}'_{1},{\bf x}'_{2},\dots,
{\bf x}'_{N}\}$. Consider the ratio
$\frac{p(\mathbb{O}|\mathbb{D}_{1})}{p(\mathbb{O}|\mathbb{D}_{2})}$. We
assume that in Step $3$ we can find the output $\mathbb{O}$ such that
the gradient of logistic regression objective w.r.t.~${\bf w}'$ is
exactly $0$. For the
classifier in Step $2$, we consider ${\bf w}'=a_1$ for the first dataset
$\mathbb{D}_{1}$ and ${\bf w}'=a_2$ for the second dataset
$\mathbb{D}_{2}$. Using the fact that
the log-likelihood of logistic regression is convex, and $a_1$ and $a_2$ are both optimal classifiers of the published data $\mathbb{O}$, thus $a_1=a_2=a$. Then, the ratio
$\frac{p(\mathbb{O}|\mathbb{D}_{1})}{p(\mathbb{O}|\mathbb{D}_{2})}$ is
computed as:
\begin{eqnarray}
\frac{p(\mathbb{O}|\mathbb{D}_{1})}{p(\mathbb{O}|\mathbb{D}_{2})}=\frac{p(\mathbb{O}|{\bf w}'=a)p({\bf w}'=a|\mathbb{D}_{1})}{p(\mathbb{O}|{\bf w}'=a)p({\bf w}'=a|\mathbb{D}_{2})}=\frac{p({\bf w}'=a|\mathbb{D}_{1})}{p({\bf w}'=a|\mathbb{D}_{2})}.\nonumber
\end{eqnarray}
Assume ${\bf w}_1=b_1$ and ${\bf w}_1=b_2$ are the optimal classifiers
for $\mathbb{D}_{1}$ and $\mathbb{D}_{2}$ after Step $1$. Therefore,
because of Laplace noise in Step $2$, $b_1+\eta_1=b_2+\eta_2=a
\Rightarrow \frac{p({\bf w}'=a|\mathbb{D}_{1})}{p({\bf
    w}'=a|\mathbb{D}_{2})}=\frac{p(\eta=\eta_1)}{p(\eta=\eta_2)}=e^{-\frac{\lambda
    N \epsilon}{2}(||\eta_1||-||\eta_2||)}$. Consequently,
$\frac{p(\mathbb{O}|\mathbb{D}_{1})}{p(\mathbb{O}|\mathbb{D}_{2})}\le
e^{\frac{\lambda N \epsilon}{2}(||b_1||-||b_2||)}\le e^{\frac{\lambda
    N \epsilon}{2}(||b_1-b_2||)}$.  The sensitivity of logistic
regression with $N$ samples and regularization parameter $\lambda$ is
atmost $\frac{2}{\lambda N}$ \cite{chaudhuri2009privacy} $\Rightarrow
\frac{p(\mathbb{O}|\mathbb{D}_{1})}{p(\mathbb{O}|\mathbb{D}_{2})}\le
e^{\epsilon}$, which completes the proof. $\qed$\\

\noindent \textbf{Proof for Proposition \ref{l:1}.} Since ${\bf w}'$ is achieved from ${\bf w}$
by adding Laplace noise, $||{\bf w}'-{\bf w}||$ is bounded. So,
$\mathbb{L}_{\lambda}({\bf w}')-\mathbb{L}_{\lambda}({\bf w})$ is
bounded using Taylor series. The rest of the proof follows from Lemma
1 in \cite{chaudhuri2009privacy}. $\qed$\\

\noindent \textbf{Proof for Proposition \ref{l:40}.} Assume there are two training datasets
$\mathbb{D}_{1}=\{{\bf x}_{1i}\}_{i=1}^N$ and $\mathbb{D}_{2}=\{{\bf
  x}_{2i}\}_{i=1}^N$, which are different at only one sample, e.g.,
without the loss of generality, we assume ${\bf x}_{1i}={\bf
  x}_{2i}$ for $i=\{1, 2,\dots,N-1\}$, and ${\bf x}_{1N}\neq {\bf
  x}_{2N}$. Assume the outputs of of Algorithm \ref{a:IPB} is
$\mathbb{O}=\{{\bf x}'_{1},{\bf x}'_{2},\dots, {\bf x}'_{N-1}, {\bf
  x}'_{N}\}$. Consider the ratio
\begin{eqnarray}
\frac{p(\mathbb{O}|\mathbb{D}_{1})}{p(\mathbb{O}|\mathbb{D}_{2})}&=&\frac{p({\bf x}'_1|{\bf x}_1)\dots p({\bf x}'_{N-1}|{\bf x}_{N-1})p({\bf x}'_{N}|{\bf x}_{1N})}{p({\bf x}'_1|{\bf x}_1)\dots p({\bf x}'_{N-1}|{\bf x}_{N-1})p({\bf x}'_{N}|{\bf x}_{2N})}\nonumber\\
&&=\frac{e^{-0.5\epsilon||{\bf x}'_{N}-{\bf x}_{1N}||}}{e^{-0.5\epsilon||{\bf x}'_{N}-{\bf x}_{2N}||}}\le e^{0.5\epsilon||{\bf x}_{1N}-{\bf x}_{2N}||} \le e^{\epsilon},\nonumber
\end{eqnarray}
where the last equation is from the fact that $||{\bf x}_i|| \le 1,
\forall i$. Thus, the input perturbation algorithm is
$\epsilon$-private. $\qed$\\

\noindent \textbf{Proof for Proposition \ref{l:3}.}
The proof is similar to \cite{chaudhuri2009privacy}. For the sake of completeness, following Lemma \ref{l:2}, define $G({\bf w})=H_2({\bf
  w})+\lambda\frac{||w||^2}{2}$ and $g=H^{'}_2({\bf w})-H_2({\bf w})$,
where $H^{'}_2({\bf w})=\frac{1}{N}\sum_{i=1}^{N} \Big[\big(y_i{\bf
    w}^T{\bf x}'_i\big) -\log(\sum_{l=1}^C e^{{\bf w}^T{\bf
      x}'_i})\Big]$ and $H_2({\bf w})=\frac{1}{N}\sum_{i=1}^{N}
\Big[\big(y_i{\bf w}^T{\bf x}_i\big) -\log(\sum_{l=1}^C e^{{\bf w}^T{\bf x}_i})\Big]$. Then, $||\nabla g||=||\frac{1}{N}\Big[\sum_{i=1}^N\Big(-(y_i-p(y_i=1|{\bf x}_i,{\bf w})\Big){\bf x}'_i
+\sum_{i=1}^N\Big(-(y_i-p(y_i=1|{\bf x}_i,{\bf w})\Big){\bf x}_i\Big]||
\le\frac{1}{N}\sum_{i=1}^N||{\bf x}'_i-{\bf x}_i|| \le (N2d\log\frac{d}{\delta})/(N\epsilon)=2d\log\frac{d}{\delta}/\epsilon$,
where the last inequality comes from the fact $||{\bf x}'_i-{\bf
  x}_i||\sim e^{\frac{-\epsilon||\eta||}{2}}$ and ${\bf x}'_i$, ${\bf
  x}_i \in \mathbb{R}^d$. Note that even though $||{\bf x}_i||$ is upper bounded by $1$ $\forall i$, $||{\bf x}'_i||$ is not upper bounded by $1$ since ${\bf x}'_i={\bf x}_i+\eta$ where $\eta \sim e^{\frac{-\epsilon||\eta||}{2}}$. Hence, $||\nabla g||$ can not be trivially upper bounded by $2$. Moreover, $v^T\nabla^2(G+g)v$ is lower
bounded by $\lambda$. Thus, $||{\bf w}'-{\bf w}||\le \frac{2 d
  \log \frac{d}{\delta}}{\lambda \epsilon}$. By Taylor expansion, $\mathbb{L}_{\lambda}({\bf
  w}')=\mathbb{L}_{\lambda}({\bf w}) +
\nabla\mathbb{L}_{\lambda}({\bf w})({\bf w}'-{\bf
  w})+\frac{1}{2}({\bf w}'-{\bf
  w})^T\nabla^2\mathbb{L}_{\lambda}({\bf w})({\bf w}'-{\bf w})
\le \mathbb{L}_{\lambda}({\bf w})+\frac{1}{2}||{\bf w}'-{\bf
  w}||^2(\lambda+1)$. This completes the proof.$\qed$\\

\section{ Acknowledgments}
This material is based upon work supported by the National Science Foundation under Grant CNS-1314956.

\bibliography{refs}

\begin{thebibliography}{}

\bibitem[\protect\citeauthoryear{Bartlett, Jordan, and
  McAuliffe}{2006}]{bartlett2006convexity}
Bartlett, P.~L.; Jordan, M.~I.; and McAuliffe, J.~D.
\newblock 2006.
\newblock Convexity, classification, and risk bounds.
\newblock {\em Journal of the American Statistical Association}
  101(473):138--156.

\bibitem[\protect\citeauthoryear{Blum, Ligett, and
  Roth}{2008}]{blum2008learning}
Blum, A.; Ligett, K.; and Roth, A.
\newblock 2008.
\newblock A learning theory approach to non-interactive database privacy.
\newblock In {\em Proceedings of the fortieth annual ACM symposium on Theory of
  computing},  609--618.

\bibitem[\protect\citeauthoryear{Chaudhuri and
  Monteleoni}{2009}]{chaudhuri2009privacy}
Chaudhuri, K., and Monteleoni, C.
\newblock 2009.
\newblock Privacy-preserving logistic regression.
\newblock In {\em Advances in neural information processing systems},
  289--296.

\bibitem[\protect\citeauthoryear{Chen \bgroup et al\mbox.\egroup
  }{2011}]{chen2011publishing}
Chen, R.; Mohammed, N.; Fung, B.~C.; Desai, B.~C.; and Xiong, L.
\newblock 2011.
\newblock Publishing set-valued data via differential privacy.
\newblock In {\em Proceedings of the International Conference on Very Large
  Data Bases}, number~11,  1087--1098.

\bibitem[\protect\citeauthoryear{Dimitrakakis \bgroup et al\mbox.\egroup
  }{2014}]{dimitrakakis2014robust}
Dimitrakakis, C.; Nelson, B.; Mitrokotsa, A.; and Rubinstein, B.
\newblock 2014.
\newblock Robust and private bayesian inference.
\newblock In {\em Proceedings of the International Conference on Algorithmic
  Learning Theory},  291--305.

\bibitem[\protect\citeauthoryear{Dwork, Roth, and
  others}{2014}]{dwork2014algorithmic}
Dwork, C.; Roth, A.; et~al.
\newblock 2014.
\newblock The algorithmic foundations of differential privacy.
\newblock {\em Foundations and Trends{\textregistered} in Theoretical Computer
  Science}  211--407.

\bibitem[\protect\citeauthoryear{Dwork}{2008}]{dwork2008differential}
Dwork, C.
\newblock 2008.
\newblock Differential privacy: A survey of results.
\newblock In {\em International Conference on Theory and Applications of Models
  of Computation},  1--19.
\newblock Springer.

\bibitem[\protect\citeauthoryear{Fung \bgroup et al\mbox.\egroup
  }{2010}]{fung2010privacy}
Fung, B.; Wang, K.; Chen, R.; and Yu, P.~S.
\newblock 2010.
\newblock Privacy-preserving data publishing: A survey of recent developments.
\newblock {\em ACM Computing Surveys (CSUR)} 42(4):14.

\bibitem[\protect\citeauthoryear{Lee and Clifton}{2014}]{lee2014top}
Lee, J., and Clifton, C.~W.
\newblock 2014.
\newblock Top-k frequent itemsets via differentially private fp-trees.
\newblock In {\em Proceedings of the International Conference on Knowledge
  Discovery and Data Mining},  931--940.

\bibitem[\protect\citeauthoryear{Lyu, Su, and Li}{2017}]{lyu2017understanding}
Lyu, M.; Su, D.; and Li, N.
\newblock 2017.
\newblock Understanding the sparse vector technique for differential privacy.
\newblock In {\em Proceedings of the International Conference on Very Large
  Data Bases},  637--648.

\bibitem[\protect\citeauthoryear{Minka}{2003}]{minkacomparison}
Minka, T.~P.
\newblock 2003.
\newblock A comparison of numerical optimizers for logistic regression.

\bibitem[\protect\citeauthoryear{Mivule}{2012}]{mivule2012utilizing}
Mivule, K.
\newblock 2012.
\newblock Utilizing noise addition for data privacy, an overview.
\newblock In {\em Proceedings of the International Conference on Information
  and Knowledge Engineering (IKE)}, ~1.

\bibitem[\protect\citeauthoryear{Mohammed \bgroup et al\mbox.\egroup
  }{2011}]{mohammed2011differentially}
Mohammed, N.; Chen, R.; Fung, B.; and Yu, P.~S.
\newblock 2011.
\newblock Differentially private data release for data mining.
\newblock In {\em Proceedings of the International Conference on Knowledge
  Discovery and Data Mining},  493--501.

\bibitem[\protect\citeauthoryear{Samarati and
  Sweeney}{1998}]{samarati1998generalizing}
Samarati, P., and Sweeney, L.
\newblock 1998.
\newblock Generalizing data to provide anonymity when disclosing information.
\newblock In {\em PODS},  188.

\bibitem[\protect\citeauthoryear{Sarwate and
  Chaudhuri}{2013}]{sarwate2013signal}
Sarwate, A.~D., and Chaudhuri, K.
\newblock 2013.
\newblock Signal processing and machine learning with differential privacy:
  Algorithms and challenges for continuous data.
\newblock {\em IEEE signal processing magazine} 30(5):86--94.

\bibitem[\protect\citeauthoryear{Vapnik and
  Vapnik}{1998}]{vapnik1998statistical}
Vapnik, V.~N., and Vapnik, V.
\newblock 1998.
\newblock {\em Statistical learning theory}, volume~1.
\newblock Wiley New York.

\bibitem[\protect\citeauthoryear{Walker and
  Duncan}{1967}]{walker1967estimation}
Walker, S.~H., and Duncan, D.~B.
\newblock 1967.
\newblock Estimation of the probability of an event as a function of several
  independent variables.
\newblock {\em Biometrika} 54:167--179.

\bibitem[\protect\citeauthoryear{Wang, Fienberg, and
  Smola}{2015}]{wang2015privacy}
Wang, Y.-X.; Fienberg, S.; and Smola, A.
\newblock 2015.
\newblock Privacy for free: Posterior sampling and stochastic gradient monte
  carlo.
\newblock In {\em Proceedings of the International Conference on Machine
  Learning},  2493--2502.

\bibitem[\protect\citeauthoryear{Xiao, Xiong, and
  Yuan}{2010}]{xiao2010differentially}
Xiao, Y.; Xiong, L.; and Yuan, C.
\newblock 2010.
\newblock Differentially private data release through multidimensional
  partitioning.
\newblock {\em Secure Data Management} 6358:150--168.

\end{thebibliography}
\bibliographystyle{aaai}

\end{document}